\documentclass[twoside,11pt]{article}

\usepackage{jmlr2e}
\usepackage{amssymb}
\usepackage{graphics}

\usepackage{subfigure} 
\usepackage{natbib}

\usepackage{algorithm}
\usepackage{algorithmic}

\usepackage{tikz}
\usepackage{array}
\usepackage{color}
\usepackage{amssymb}
\usepackage{amsmath}
\usepackage{amsfonts}

\usepackage{adjustbox}

\newcolumntype{R}[2]{%
    >{\adjustbox{angle=#1,lap=\width-(#2)}\bgroup}%
    l%
    <{\egroup}%
}

\usepackage{array}

\usetikzlibrary{fit,positioning, shapes.geometric, arrows}

\tikzstyle{startstop} = [rectangle, rounded corners, minimum width=1.3cm, minimum height=0.8cm,text centered,
draw=black, fill=red!30]
\tikzstyle{observedData} = [rectangle, minimum width=2.5cm, minimum height=0.7cm, text centered, draw=black, 
fill=orange!30]
\tikzstyle{latent} = [rectangle, rounded corners, minimum width=1.7cm, minimum height=7mm, text centered, 
draw=black, fill=blue!30]
\tikzstyle{decision} = [diamond, minimum width=3cm, minimum height=1cm, text centered, draw=black, 
fill=green!30]
\tikzstyle{arrow} = [thick,->,>=stealth]

\usepackage{url}

\jmlrheading{1}{2000}{1-48}{4/00}{10/00}{Jussi Gillberg, Pekka Marttinen and Samuel Kaski}

\ShortHeadings{Latent noise}{Gillberg et al.}
\firstpageno{1}

\begin{document}

\title{Multiple Output Regression with Latent Noise}

\author{\name Jussi Gillberg \email jussi.gillberg@aalto.fi \\
       \name Pekka Marttinen \email pekka.marttinen@aalto.fi \\
	\addr Helsinki Institute for Information Technology HIIT \\ Department of Computer
Science \\ PO Box 15600, Aalto University, 00076 Aalto, Finland 
       \AND
       \name Matti Pirinen \email matti.pirinen@helsinki.fi  \\
	\addr Institute for Molecular Medicine Finland (FIMM) \\ University of Helsinki, Finland 
       \AND
       \name Antti J. Kangas \email antti.kangas@computationalmedicine.fi‎  \\	
       \name Pasi Soininen \thanks{PS and MAK are also at NMR Metabolomics Laboratory,School of
Pharmacy,University of Eastern Finland,Kuopio,Finland; MRJ is also at Center for Life Course Epidemiology and Systems Medicine, University
of Oulu, Finland and Biocenter Oulu, University of Oulu, Finland and Unit of Primary Care, Oulu
University Hospital, Oulu, Finland; MAK is also at Computational Medicine, School of Social and Community
Medicine and the Medical Research Council Integrative Epidemiology Unit, University of Bristol, Bristol,
UK} \email pasi.soininen@computationalmedicine.fi \\
	\addr Computational Medicine \\ Center for Life Course Epidemiology and Systems Medicine \\ University of Oulu and Oulu University
Hospital, Oulu,Finland 
       \AND
       \name Mehreen Ali \email mehreen.ali@aalto.fi  \\
	\addr Helsinki Institute for Information Technology HIIT \\ Department of Computer
Science \\ PO Box 15600, Aalto University, 00076 Aalto, Finland 
       \AND
       \name Aki S. Havulinna \email aki.havulinna@thl.fi \\
	\addr Department of Health \\ National Institute for Health and Welfare, 
Helsinki, Finland 
       \AND
       \name Marjo-Riitta J\"arvelin$^{\ast}$ \email m.jarvelin@imperial.ac.uk‎  \\
	\addr Department of Epidemiology and Biostatistics \\ MRC Public Health England, PHE, Centre for
Environment and Health, School of Public Health \\ Imperial College London, United Kingdom
       \AND
       \name Mika Ala-Korpela$^{\ast}$ \email mika.ala-korpela@computationalmedicine.fi
\\
	\addr Computational Medicine \\ Center for Life Course Epidemiology and Systems Medicine \\ University of Oulu and Oulu University
Hospital, Oulu,Finland
       \AND
       \name Samuel Kaski$^{\ast}$ \email samuel.kaski@aalto.fi \\
	\addr Helsinki Institute for Information Technology HIIT \\ Department of Computer
Science \\ PO Box 15600, Aalto University, 00076 Aalto, Finland}

\editor{}

\maketitle

\begin{abstract}%
In high-dimensional data, structured noise caused by observed and
unobserved factors affecting multiple target variables simultaneously,
imposes a serious challenge for modeling, by masking the often weak
signal. Therefore, (1) explaining away the structured noise in
multiple-output regression is of paramount importance. Additionally,
(2) assumptions about the correlation structure of the regression
weights are needed. We note that both can be formulated in a natural
way in a latent variable model, in which both the interesting signal and
the noise are mediated through the same latent factors. Under this assumption, the signal
model then borrows strength from the noise model by encouraging
similar effects on correlated targets. We introduce a hyperparameter
for the \emph{latent signal-to-noise ratio} which turns out to be
important for modelling weak signals, and an ordered
infinite-dimensional shrinkage prior that resolves the rotational
unidentifiability in reduced-rank regression models. Simulations and prediction experiments with metabolite, 
gene expression, FMRI measurement, and macroeconomic time series data show that our model equals or exceeds 
the state-of-the-art performance and, in particular, outperforms the standard approach 
of assuming independent noise and signal models.
\end{abstract}

\begin{keywords}
Bayesian reduced-rank regression, latent variable models, latent signal-to-noise ratio, multiple-output regression,  
nonparametric Bayes, shrinkage priors, structured noise, weak effects
\end{keywords}

\section{Introduction} 

Explaining away structured noise is one of the cornerstones for successful modeling of high-dimensional output
data in the regression framework 
\citep{Fusi12,klami2013bayesian,rai2012,rakitsch2013all,stegle2012using,Virtanen11}. The
structured noise refers to dependencies between response variables, which are unrelated to the dependencies 
of interest between the response variables and the covariates. It is noise caused by observed and unobserved confounders that affect multiple variables 
simultaneously. Common observed confounders in medical and biological data include age and sex of an 
individual, whereas unobserved confounders include, for example, the state of the cell being measured, 
measurement artefacts influencing multiple probes, or other unrecorded experimental conditions. When not 
accounted for, structured noise may both hide interesting relationships and result in spurious findings 
\citep{leek2007capturing,kang2008accurate}.

The effects of known confounders can be removed straightforwardly by using supervised methods. For the
unobserved confounders, a routinely used approach for explaining away structured noise has been to assume
\textit{a priori} independent effects for the interesting and uninteresting factors. For example, in the
factor regression setup \citep{bernardo2003bayesian,stegle2010bayesian,Fusi12}, the target variables $Y$ are
assumed to have been generated as
\begin{equation}
 Y = X \mathbf{\Theta} + H \Lambda + E, \label{eq:factor_regression_modeling}
\end{equation} 
where $Y_{N \times K}$ is the matrix of $K$ target variables (or dependent variables) and $X_{N \times P}$
contains the covariates (or independent variables), for the $N$ observations. The model parameter matrix $H_{N
\times S_2}$ comprises the unknown latent factors and $\Lambda_{S_2 \times K}$ the factor loadings, which are
used to model away the structured noise. The term $E_{N \times K}$ represents independent 
unstructured noise and the elements of $E$ are independently distributed, $\text{vec}(E) \sim 
\mathcal{N}(0,I_{NK})$. In this paper we call this model \textbf{independent-noise BRRR}. To reduce the effective 
number of parameters in the regression coefficient matrix $\mathbf{\Theta}_{P \times
K}$, a low-rank structure may be assumed:
\begin{equation}
 \mathbf{\Theta} = \Psi \; \Gamma, \label{eq:low_rank}
\end{equation}
where the rank $S_{1}$ of parameters $\Psi_{P \times S_{1}}$ and $\Gamma_{S_{1} \times K}$ is substantially
lower than the number of target variables $K$ and covariates $P$. The low-rank decomposition of the regression
coefficient matrix (\ref{eq:low_rank}) may be given an interpretation
whereby the covariates $X$ affect $S_1$
latent components with coefficients specified in $\Psi$, and the components, in turn, affect the target $Y$
with coefficients $\Gamma$.

\begin{figure}[h!]
\centering
\begin{tabular}{ccc}

\begin{tikzpicture}[node distance=2cm]

\node (X) [observedData] {Covariates};
\node (Z) [latent, below of=X,yshift=0.4cm] {Latent space};
\node (Y) [observedData, below right of=Z, yshift=-0.3cm] {Target variables};
\node (noise) [startstop, right of=Z, xshift=0.5cm] {Noise};

\draw [arrow] (X) -- (Z);
\draw [arrow] (Z) -- node[anchor=east] {structured effect} (Y);
\draw [arrow] (noise) -- node[anchor=west] {structured noise} (Y);

\end{tikzpicture}

& \hspace{2mm} &
 
\begin{tikzpicture}[node distance=2cm]

\node (X) [observedData] {Covariates};
\node (Z) [latent, below right of=X, yshift=-2mm] {Latent space};
\node (Y) [observedData, below of=Z, yshift=0.3cm] {Target variables};
\node (noise) [startstop, right of=X, xshift=0.7cm] {Noise};

\draw [arrow] (X) -- (Z);
\draw [arrow] (Z) --node[align=center, anchor=west] {joint structured \\[1mm] effect \& noise} (Y);
\draw [arrow] (noise) -- (Z);

\end{tikzpicture} \\
(a) & & (b) \\
\end{tabular}
\caption{Illustration of (a) \emph{a priori} independent interesting and uninteresting effects and (b) the
latent noise assumption. Latent noise is mediated to the target variable measurements through a common
subspace with the interesting effects.}
\label{Fig_illustraatio}
\end{figure}
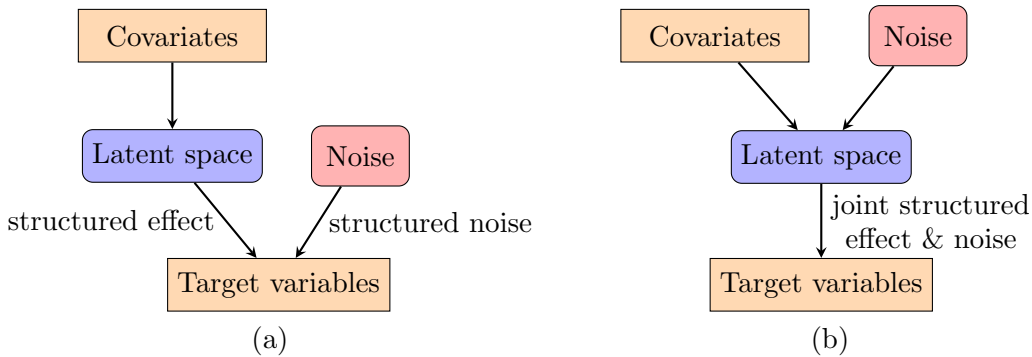

Another line of work in multiple output prediction has focused on borrowing information from the correlation
structure of the target variables when learning the regression model. The intuition stems from the observation
that correlated targets are often seen to be affected similarly by the covariates, for example in
genetic applications \citep[see, e.g.,][]{davis2014correlation,inouye2012novel}. One popular method, GFlasso
\citep{kim2009multivariate}, learns the regression coefficients using
\begin{align}
\nonumber\hat{\Theta}& = \text{argmin} \sum_k (\mathbf{y_k}-X\mathbf{\theta_{k}})^T
(\mathbf{y_k}-X\mathbf{\theta_k}) +
\\
& \lambda \sum_j \sum_k |\theta_{jk}| + \gamma\sum_{(m,l)\in
E}r_{ml}^2\sum_j|\theta_{jm}-\text{sign}(r_{ml})\theta_{jl}|,
\label{eq:gflasso_objective}
\end{align}
where the $\mathbf{\theta_k}$ are the columns of $\hat{\Theta}$. Two regularization parameters are introduced:
$\lambda$ represents the standard Lasso penalty, and $\gamma$ encourages the effects $\theta_{jm}$ and
$\theta_{jl}$ of the $j$th covariate on correlated outputs $m$ and $l$
to be similar. Here $r_{ml}$
represents the correlation between the $m$th and $l$th phenotypes. The
$E$ is an \textit{a priori} specified
correlation graph for the output variables, with edges representing
correlations to be accounted for in the model.

In this paper we propose a model that simultaneously learns the
structured noise and encourages the sharing of information between the
noise and the regression models. To motivate the new model, we note
that by assuming independent prior distributions on $\Gamma$ and
$\Lambda$ in model (\ref{eq:factor_regression_modeling}), one
implicitly assumes independence of the interesting and uninteresting
effects, caused by covariates $X$ and unknown factors $H$,
respectively (Fig. \ref{Fig_illustraatio}a). The assumption is appealing for example when 
explaining away batch effects \citep{Fusi12} in high-dimensional data, but may be inadequate in the presence of other types of noise in molecular biology, where gene expression and metabolomics measurements record concentrations of compounds generated
by ongoing latent biological processes. In this kind of situations, a limited set of covariates, such as single nucleotide
polymorphisms (SNPs),  determines the activity of the latent 
process only partially and all other activity of the process
 is due to  unrecorded factors. In such cases, the noise
affects the measurement levels through the very same process as the
interesting signal (Fig. \ref{Fig_illustraatio}b), and rather than
assuming independence of the effects, an assumption about parallel
effects would be more appropriate. We refer to this type of noise as
\emph{latent noise} as it can be considered to affect the same latent
subspace as the interesting effects. We note that in practice both types of
structured noise are likely to be present. In this work, our main focus is on the latent noise, but we also present a comparison with a model that includes both types of structured noise simultaneously.

\begin{figure}[ht!]
\centering
\begin{tabular}{ccc}
. \includegraphics[width=0.3\textwidth]{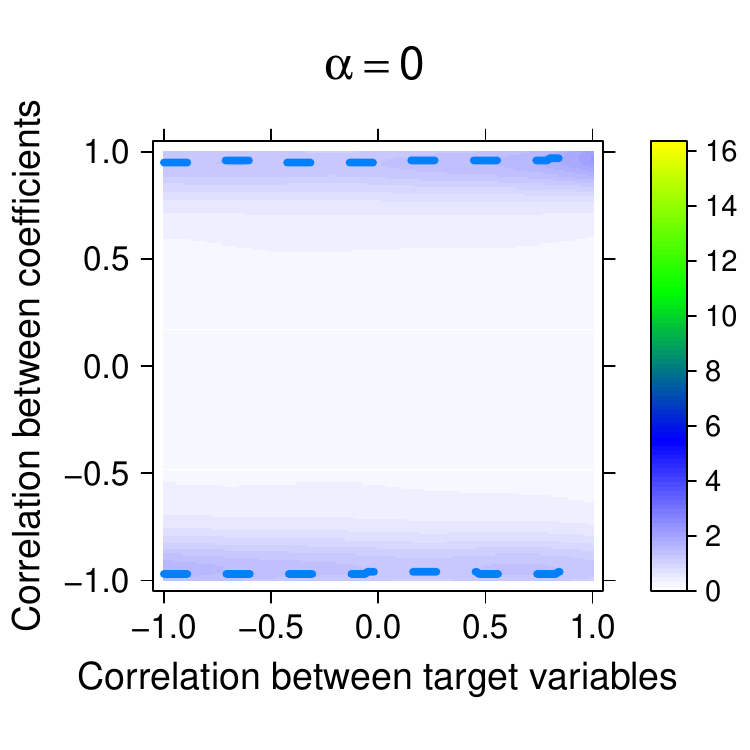} 
&
 \includegraphics[width=0.3\textwidth]{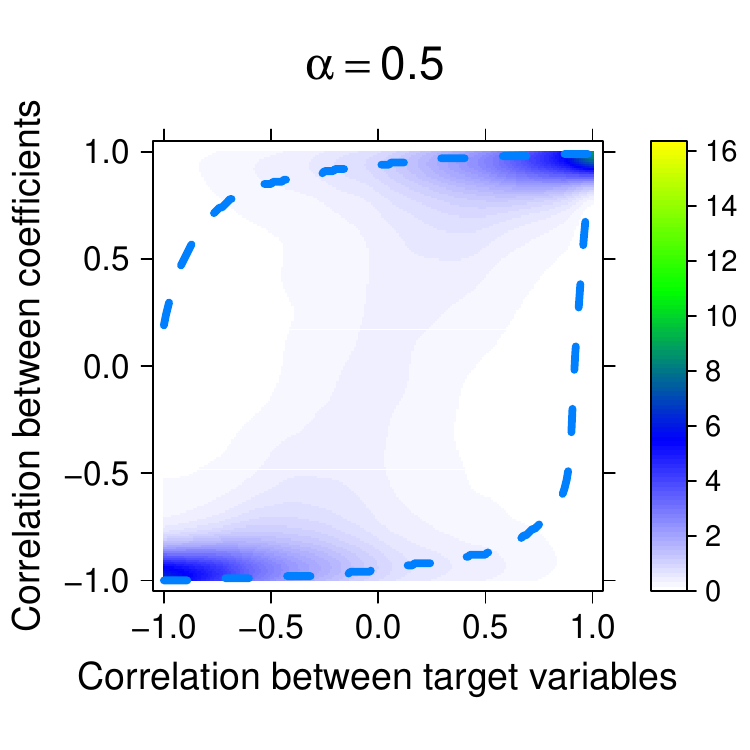} 
&
 \includegraphics[width=0.3\textwidth]{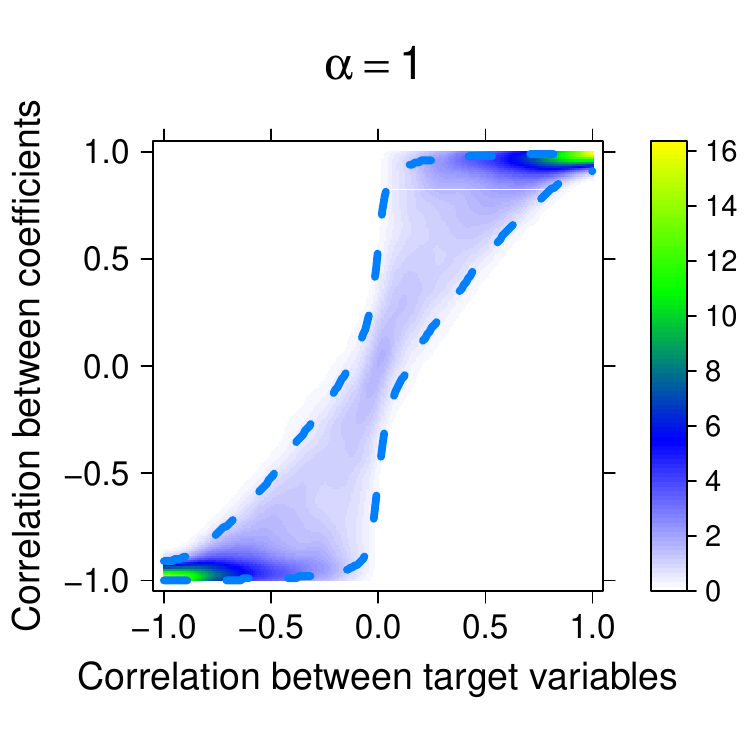} \\
(a) & (b) & (c) \\
\end{tabular}
\caption{Conditional distribution of the correlation between
  regression coefficients, given the correlation between the
  corresponding target variables. In (a) the model
  (\ref{eq:factor_regression_modeling}) assumes \emph{a priori}
  independent regression and noise models, and in (c) the model
  (\ref{eq:new_model}) makes the latent noise assumption. (b) A mixture
  of the models in a and c. The data were generated using equation
  (\ref{eq:mixture_model}), as described in Section
  \ref{sec:simulation_exp}, and $\alpha$ denotes the relative
  proportion of latent noise in data generation. The dashed lines
  denote the 95\% confidence intervals of the conditional
  distributions.}
\label{Fig_ehdolliset_korrelaatiot}
\end{figure}

A natural way to encode the assumption of latent noise is to use the following model structure:
\begin{equation}
 Y = (X \Psi + \Omega) \; \Gamma  +  E, 
\label{eq:new_model}
\end{equation} 
where the $\Omega_{N \times S_{1}}$ is a matrix consisting of unknown
latent factors. In (\ref{eq:new_model}), $\Gamma$ mediates the effects
of both the interesting and uninteresting signals on the target
variables. We note that the change required in the model structure is
small, and has in fact been presented earlier
(\citealt{bo2009supervised}; recently extended with an Indian Buffet
Process prior on the latent space by \citealt{bargi2014non}). We now
proceed to using the structure (\ref{eq:new_model}) for GFlasso-type
sharing of information (\ref{eq:gflasso_objective}) between the
regression and noise models while simultaneously explaining away
structured noise. To see that the information sharing between noise
and regression models follows immediately from model
(\ref{eq:new_model}), one can consider simulations generated from the
model. The \emph{a priori} independence assumption of model
(\ref{eq:factor_regression_modeling}) results in uncorrelated
regression weights regardless of the correlations between target
variables (Figure \ref{Fig_ehdolliset_korrelaatiot}a). The assumption
of latent noise (\ref{eq:new_model}), however, encourages the
regression weights to be correlated in a similar way as the target
variables are (Figure \ref{Fig_ehdolliset_korrelaatiot}c).

In this work, we focus on modelling weak signals in high-dimensional data with structured
noise, where we consider effects that explain a tiny portion, say
$<1\%$, of the variance of the target variables as weak. We have hypothesized above that a model with the structure
(\ref{eq:new_model}) might be particularly well-suited for this purpose.
Additionally, (\textit{i}) particular emphasis must be put on defining
adequate prior distributions to distinguish the weak effects from
noise as effectively as possible, and (\textit{ii}) scalability to
large sample size is needed in order to have any chance of learning the weak
effects. For (\textit{i}), we define \emph{latent signal-to-noise
  ratio} $\beta$ as a generalization of the standard signal-to-noise
ratio in the latent space:
\begin{equation}
 \beta =  \frac{\text{Trace}( \text{Var}(X \; \Psi))}{\text{Trace}( \text{Var}(\Omega)) },
\label{eq:latent_SNR}
\end{equation}
We use the latent signal-to-noise ratio as a hyperparameter in our model, and show that it is a key parameter affecting model
performance. It can be either learned or set using prior knowledge.
In addition, we introduce an ordered infinite-dimensional shrinkage
prior that resolves the inherent rotational ambiguity in the model
(\ref{eq:new_model}), by sorting both signal and noise components by
their importance.  Finally, we present efficient inference methods for
the model.

\section{Related work}

Simultaneously solving multiple real-valued prediction tasks with the same set of covariates is called multiple-output 
regression \citep{breiman1997predicting}; and more generally sharing of statistical strength between related 
tasks is called multitask learning \citep{Baxter96,caruana1997multitask}. The data consist of $N$ input-output 
pairs $(\mathbf{x}_n,\mathbf{y}_n)_{n=1,\ldots,N}$; the $P$-dimensional input vectors $\mathbf{x}$  
(covariates) are used for predicting $K$-dimensional vectors $\mathbf{y}$ of target variables. The common 
approach to dealing with structured noise due to unobserved confounders is to apply factor regression modeling 
(\ref{eq:factor_regression_modeling}) \citep{bernardo2003bayesian} and to explain away the structured noise 
using a noise model that is assumed to be \emph{a priori} independent of the regression model 
\citep{stegle2010bayesian,Fusi12,rai2012,Virtanen11,klami2013bayesian,rakitsch2013all}. A recent Bayesian 
reduced-rank regression (BRRR) model \citep{marttinen2014assessing} implements the routine assumption of the 
independence of the regression and noise models; we will include it in the comparison studies of this paper.

Methods for multiple-output regression without the structured noise model have been proposed in other fields. 
In the application fields of genomic selection and multi-trait quantitative trait loci mapping, solutions 
\citep{Yi09,Xu08,Calus11,stephens2013unified} for low-dimensional target variable vectors ($K < 10$) have been 
proposed, but these methods do not scale up to the currently emerging needs of analyzing higher-dimensional 
target variable data. Additionally, sparse multiple-output regression models have been proposed for prediction 
of phenotypes from genomic data \citep{kim2009multivariate,sohn2012joint}.

Many methods for multi-task learning have been proposed in the field of kernel methods 
\citep{evgeniou2007multi}. These methods do not, however, scale up to data sets with several thousands of 
samples, required for predicting the weak effects. Other relevant work include a recent method based on the 
BRRR presented by \cite{foygel2012}, but it does not scale to the dimensionalities of our experiments 
either. Methods for high-dimensional phenotypes have been proposed in the field of expression quantitative 
trait loci mapping \citep{bottolo2011bayesian} for the related task of finding associations (and avoiding false 
positives) rather than prediction, which is our main focus. Also functional assumptions 
\citep{wang2012integrative} have been used to constrain related learning problems.

\section{Model}
In this Section, we present the details of our new model, Bayesian reduced rank regression with latent noise (‘latent-noise BRRR’), show how the
hyperparameters can be set using the latent signal-to-noise ratio, and
analyze theoretically some properties of the infinite-dimensional shrinkage prior.

\subsection{Model details: latent-noise BRRR} \label{sec_latent_noise_brrr}
\begin{figure}[tb]
\centering
\begin{tikzpicture}
\tikzstyle{main}=[circle, minimum size = 5mm, thick, draw =black!80, node distance = 14mm]
\tikzstyle{hyper2}=[draw=none, fill=none]
\tikzstyle{connect}=[-latex, thick]
\tikzstyle{box}=[rectangle, draw=black!100]
  \node[main, fill = black!100] (Y) [label=below:$Y_{N\times K}$] { };
  \node[main] (Psi) [above left=of Y, label=left:$\Psi_{P\times S_1}$] { };
  \node[main, fill = black!100] (X) [below left=of Y, label=below:$X_{N\times P}$] { };
  \node[main, fill = green!40] (Gamma) [above right=of Y, label=right:$\Gamma_{S_1 \times K}$] { };
  \node[main] (Gamma_shrink) [above right=of Gamma, label=right:{$\Phi_{S_1\times K}^{\Gamma}$}] { };
  \node[main] (delta_star) [above right=of Psi, label=left:${\delta_l^\ast, l=1,\ldots,S_1}$] { };
  \node[hyper2] (a3a4) [above=of delta_star] {$a_1,a_2$};
  \node[hyper2] (nu) [above=of Gamma_shrink] {$\nu$};
  \node[main, fill = black!20] (sigmas) [below right=of Gamma, label=below:{$\sigma_j^2, j=1,\ldots,K$}] { };
  \node[main, fill = black!20] (H) [below right=of Y, label=below:{$\Omega_{N\times S_1}$}] { };
  \path (X) edge [connect] (Y)
  	(Psi) edge [connect] (Y)
  	(Gamma) edge [connect] (Y)
  	(delta_star) edge [connect] (Psi)
	(Gamma_shrink) edge [connect] (Gamma)
  	(delta_star) edge [connect] (Gamma)
		(a3a4) edge [connect] (delta_star)
		(sigmas) edge [connect] (Y)
		(H) edge [connect] (Y)
		(nu) edge [connect] (Gamma_shrink);
\end{tikzpicture}
\caption{Graphical representation of latent-noise BRRR. The observed
  data are denoted by black circles, variables related to the
  reduced-rank regression part of the model by white circles,
  variables related only to the noise model are denoted by gray
  circles, and variables related to both the regression and the
  structured noise model are denoted with green circles. The matrix
  $\Phi_{S_1\times K}^{\Gamma}$ comprises the sparsity parameters for
  the $K$ target variables for the components.}
\label{fig:dag}
\end{figure}
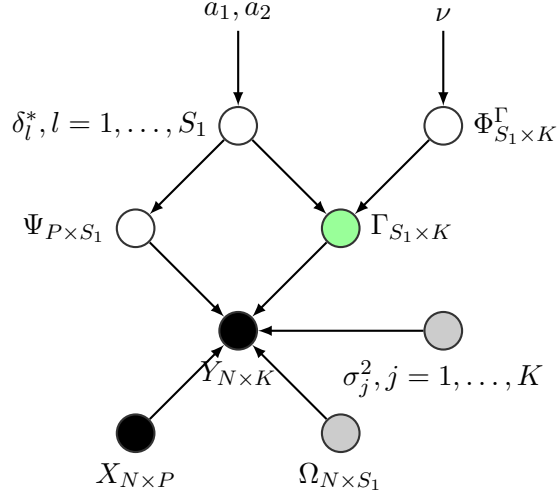

Our model is given by 
\begin{equation}
Y=(X\Psi + \Omega)\Gamma+E,\label{eq:brr_model}%
\end{equation}
where $Y_{N\times K}$ contains the $K$-dimensional response variables for $N$ observations, and $X_{N\times P}$
contains the predictor variables. The product $\Theta=\Psi\Gamma$, of $\Psi_{P\times S_{1}}$ and
$\Gamma_{S_{1}\times K}$, results in a regression coefficient matrix with rank $S_{1}$. The $\Omega_{N\times
S_{1}}$ contains unknown latent factors representing the latent noise. Finally, $E_{N\times K}=[e_{1},\ldots,e_{N}]^{T},$ with $e_{i}\sim N(0,\Sigma),$ where $\Sigma=diag(\sigma_{1}^{2},\ldots,\sigma_{K}^{2})$ is a matrix of uncorrelated target variable-specific
noise vectors. Figure~\ref{fig:dag} displays graphically the structure of the model. In the figure, the node corresponding to the parameter $\Gamma$ that is shared by the regression and noise models is highlighted with green.

Similarly to a recent BRRR model \citep{marttinen2014assessing} and the Bayesian infinite sparse factor analysis model \citep{bhattacharya2011sparse}, we assume the 
number of components $S_{1}$ connecting the covariates to the targets to be infinite. Accordingly, the number of rows in the weight matrix $\Gamma$, and the 
numbers of columns in $\Psi$ and $\Omega$, are infinite. The 
low-rank nature of the model is enforced by shrinking the columns of $\Psi$ and rows of $\Gamma$ and $\Omega$ increasingly with the growing column/row index, such that only a small number of columns/rows are influential in practice. The 
increasing shrinkage also solves any rotational unidentifiability issues by enforcing the model to mediate the strongest effects through the first columns/rows. In Section \ref{sec:validity} we explore the basic properties of the infinite-dimensional prior, to ensure its soundness.
The hierarchical priors for the projection weight matrix $\Gamma$, where $\Gamma = [\gamma_{hj}]$, are set as follows:
\begin{gather}
\gamma_{hj}|\phi_{hj}^\Gamma,\tau_{h}\sim N\left(  0,\left(  \phi_{hj}^\Gamma\tau
_{h}\right)  ^{-1}\right), \hspace{1mm}\phi_{hj}^\Gamma\sim\text{Ga}(\nu/2,\nu/2),\nonumber\\ 
\tau_{h}=\prod_{l=1}^{h}\delta_{l}, \quad
\delta_{1}\sim\text{Ga}(a_{1},1),\quad\delta_{l}\sim\text{Ga}(a_{2},1),\quad
l\geq2.\quad \label{eq:Gamma_prior}%
\end{gather}
Here $\tau_{h}$ is a global shrinkage parameter for the $h$th row of $\Gamma$ and the $\phi_{hj}^\Gamma$s are local shrinkage parameters for the individual elements of $\Gamma$, to provide additional flexibility over the global shrinkage priors. The same parameters $\tau_{h}$ are used to shrink the columns of the matrices $\Psi=\left[\psi_{jh}\right]$ and $\Omega=\left[\omega_{jh}\right]$, because the scales of $\Gamma$ and $\Psi$ (or $\Omega$) are not identifiable separately:
\begin{equation*}
\psi_{jh}|\tau_{h} \sim N\left(  0,\left(  \tau_{h}\right)^{-1}\right), \quad \text{and} \quad
\omega_{jh}|\tau_{h} \sim N\left(  0,\sigma_{\Omega}^2\left(  \tau_{h}\right)^{-1}\right),
\end{equation*}
where $\sigma_{\Omega}^2$ is a parameter that specifies the amount of latent noise, which is used to
regularize the model (see the next Section).
With the priors specified, the hidden factors $\Omega$ can be integrated out analytically, yielding%
\begin{equation}
y_{i}\sim N\left((\Psi\Gamma)^{T}x_{i}, \sigma_{\Omega}^2(\Gamma^*)^{T}(\Gamma^*)+\Sigma\right),\quad i=1,\ldots,N,
\label{eq:integrate_over_Omega} 
\end{equation}
where $\Gamma^*$ is obtained from $\Gamma$ by multiplying the rows of $\Gamma$ with the shrinkages $(\tau_h)^{-1/2}$ of the columns of $\Omega$.

Finally, conjugate prior distributions
\begin{gather}
 \sigma_{j}^{-2}\sim\text{Ga}(a_{\sigma},b_{\sigma}),\quad j=1,\ldots
,K,\label{eq:var.terms}%
\end{gather}
are placed on the noise parameters of the target variables.

\subsection{Regularization of latent-noise BRRR through the variance of $\Omega$} \label{sec:selecting_var}

The latent signal-to-noise ratio $\beta$ in Equation (\ref{eq:latent_SNR}) has an intuitive
interpretation: given our prior distributions for $\Psi$ and $\Omega$, the prior latent SNR indicates 
the extent to which we believe the noise to explain variation in $Y$, as compared to the variance explained by the
covariates $X$. Thus, the latent SNR acts as a regularization parameter: when the latent variables $\Omega$ are 
allowed to have a large variance, the data will be explained by the noise model rather than the covariates. 
We note that this approach to regularization is non-standard and it may have favourable characteristics compared 
to the commonly used L1/L2 regularization of regression weights. First of all, the regression weights remain 
relatively unbiased as they need not be enforced to zero to control for overfitting. This is important when the 
effects are weak: if the effects were shrunk towards zero, they might be lost completely.

Secondly, while regularizing with the \emph{a priori} selected latent SNR, the regularization parameter itself 
remains interpretable: every value of the variance parameter of $\Omega$ can be immediately interpreted as the 
percentage of variance explained by the noise model as compared to the covariates. In our experiments, we use 
cross-validation to select the variance of $\Omega$ and the interpretability of the parameter makes it 
easy to express beliefs of the plausible values based on prior knowledge. Making similar educated guesses for L1/L2
regularization parameters is not straightforward.

\subsection{Difference between latent-noise BRRR and independent-noise BRRR} 
\label{sec:latent_independent_differences}
We call the standard Bayesian reduced rank regression (Equation \ref{eq:factor_regression_modeling}), which assumes 
independent noise and signal models, the \emph{independent-noise BRRR}. The new latent-noise BRRR differs from it in
two ways: in the latent-noise BRRR
\begin{enumerate}
 \item the structure of the model is different in that the noise model uses the same projection parameters as 
the regression model, and
 \item the model is regularized by modifying the variance of the noise model. This is achieved by learning the 
latent signal-to-noise ratio parameter $\beta$.
\end{enumerate}
In Section \ref{sec:NFBC_detailed_res} we show that both of these improvements are needed to reach the 
performance differences observed. 

We emphasize that although the technical difference between the two models is minor, the models are very different 
from the conceptual point of view, as discussed in the Introduction, as well as from the practical point of 
view. In particular, it has been reported before that with weak effects the independent-noise BRRR may suffer 
from severe instability, resulting from a highly multi-modal posterior distribution and, consequently, poor 
convergence and mixing properties of the learning algorithms \citep{koop2006bayesian,marttinen2014assessing}. 
In Section \ref{sec:efficiency_of_the_algorithm}, we demonstrate how the latent noise assumption provides just 
the required additional regularization to make the formal Bayesian inference tractable even with weak effects.

As both independent structured noise and latent noise could be present, a logical extension to the 
models presented so far is to consider both noise types simultaneously,
\begin{equation}
 Y = (X \Psi + \Omega) \; \Gamma  +  H \Lambda + E,
\label{eq:latent+independent_noise_model}
\end{equation} 
where the distributional assumptions for $\Psi, \Omega$ and $\Gamma$ are the same as in latent-noise BRRR, and 
for $H $ and $\Lambda$ they follow independent-noise BRRR. The Gibbs updates for this model are straightforward 
modifications of those for the 
latent-noise BRRR and independent-noise BRRR. We have implemented also this model and study its performance in 
Section \ref{sec:latent+independent_BRRR}.

We note that both the latent-noise model is, in principle, able to express data 
generated by the independent-noise BRRR  model, and vice versa. The latent-noise BRRR model may learn noise 
components that are independent from the signal in practice, having negligible contribution from the regression 
part $X\Psi$. On the other hand, nothing prevents the independent noise model to learn some correlated regression 
and noise components. Therefore, the family of models defined by Equation (\ref{eq:latent+independent_noise_model}) that 
simultaneously includes both kinds of structured noise may have redundancy in its parameters. Indeed, the 
experiments in Section \ref{sec:latent+independent_BRRR} demonstrate only minor improvements from this model.

\subsection{Proofs of the soundness of the infinite prior} \label{sec:validity}

In this Section we verify the sensibility of the infinite
non-parametric prior, which we introduce for ordering the components
according to decreasing importance, and of a computational
approximation resulting from truncation of the infinite model.

It has been proven that in Bayesian factor models $a_1>2$ and $a_2>3$ (in our case defined in eqn \ref{eq:Gamma_prior}) is sufficient for the elements of $\Lambda\Lambda^T$ to have
finite variance in a Bayesian factor model (\ref{eq:factor_regression_modeling}), even if an infinite number of
columns with a prior similar to our model is assumed for $\Lambda$ \citep{bhattacharya2011sparse}. In this
Section we present similar characteristics for the infinite reduced-rank regression model. The detailed proofs can be found in the Supplementary material. First, in analogy to
the infinite Bayesian factor analysis model, we show that
\begin{equation}
a_1>2 \quad \text{and} \quad a_2>3
\label{eq:sufficient}
\end{equation}
is sufficient for the prediction of any of the response variables to have finite variance under the prior
distribution (Proposition 1). Second, we show that the underestimation of uncertainty (variance) resulting
from using a finite rank approximation to the infinite reduced-rank regression model decays exponentially with
the rank of the approximation (Proposition 2). For notational clarity, let $\Psi_{h}$ denote 
the $h^{\text{th}}$ column of the $\Psi$ matrix in the following. With this notation, the prediction for the $i$th response
variable can be written as
\begin{align*}
\widetilde{y_{i}}  & =x^{T}\Theta_{i}\\
& =x^{T}\sum_{h=1}^{\infty}\Psi_{h}\gamma_{hi}.
\end{align*}
Furthermore, let $\Gamma(\cdot)$ denote below the gamma function (not to be confused with the matrix $\Gamma$ used
in all other Sections of this paper).

\noindent\textbf{Proposition 1: Finite variance of predictions} Suppose that $a_{1}>2$ and $a_{2}>3$.
Then
\begin{equation}
\text{Var}(\widetilde{y_{i}})=\frac{\nu}{\nu-2}\sum_{j=1}^{P}\text{Var}%
(x_{j})\frac{\Gamma(a_{1}-2)/\Gamma(a_{1})}{1-\Gamma(a_{2}-2)/\Gamma(a_{2})}.
\label{eq:exp_ptve}%
\end{equation}
A detailed proof is provided in the Supplementary material.

\noindent\textbf{Proposition 2: Truncation error of the finite rank approximation} Let
$\widetilde{y_{i}}^{S_{1}}$ denote the prediction for the $i$th target variable when using an approximation for
$\Psi$ and
$\Gamma$ consisting of the first $S_{1}$ columns or rows only, respectively.
Then,
\[
\frac{\text{Var}(\widetilde{y_{i}})-\text{Var}(\widetilde{y_{i}}^{S_{1}}%
)}{\text{Var}(\widetilde{y_{i}})}=\left[  \frac{\Gamma(a_{2}-2)}{\Gamma
(a_{2})}\right]  ^{S_{1}},
\]
that is, the reduction in the variance of the prediction resulting from using
the approximation, relative to the infinite model, decays exponentially with
the rank of the approximation. A detailed proof is provided in the Supplementary material.

\section{Efficient computation by reparameterization} \label{sec:computation}

For estimating the parameters of the latent-noise BRRR, we use Gibbs sampling, updating the parameters
one by one by sampling them from their conditional posterior probability distributions, given the current
values of all other parameters. The bottleneck of the computation is in updating the matrix $\Psi$, and below we present a novel efficient update for this parameter.

\subsection*{Update of $\Gamma$}
The conditional distribution of the parameter matrix $\Gamma$ of latent-noise BRRR can be updated using a standard result for Bayesian linear models \citep{bishop2006pattern} which states that if
\begin{equation}
 \beta \sim N(0, \Sigma_{\beta}), \quad \text{and} \quad y|X^{\ast},\beta \sim N(X^{\ast}\beta, \Sigma_{y}),
 \label{eq:linear_model}
\end{equation}
then
\begin{equation}
 \beta|y,X^{\ast} \sim N(\Sigma_{\beta|Y}(X^{\ast T}\Sigma_{y}^{-1}y), \Sigma_{\beta|y}),
 \label{eq:posterior_mean}
\end{equation}
where
\begin{equation}
 \Sigma_{\beta|y} = (\Sigma_{\beta}^{-1} + X^{\ast T}\Sigma_y^{-1}X^{\ast})^{-1}.
\label{eq:posterior_cov}
\end{equation}
Because in our model (\ref{eq:brr_model}) the columns $E_i$ of the noise matrix are assumed independent with variances $\sigma_1^2,\ldots,\sigma_K^2$, we get
\begin{equation}
Y_i \sim N((X\Psi + \Omega)\Gamma_i,\sigma_i^2 I_N).
\end{equation}
Thus, by substituting
\begin{equation*}
 X^{\ast} \leftarrow X\Psi + \Omega, \quad \beta \leftarrow \Gamma_i, \quad \text{and} \quad \Sigma_y \leftarrow \sigma_i^2 I_N \\
\end{equation*}
into (\ref{eq:linear_model}), together with prior covariance $\Sigma_{\beta}$ derived from (\ref{eq:Gamma_prior}), we immediately obtain the posterior of $\Gamma_i$ from (\ref{eq:posterior_mean}) and (\ref{eq:posterior_cov}).

\subsection*{Updates of $\Phi^{\Gamma}, \delta, \sigma$ and $\Omega$}
The updates of the hyperparameters are the same as in Bayesian Reduced
Rank Regression, and the conditional posterior distributions of the
hyperparameters can be found in the Supplementary material of
\citet{marttinen2014assessing}. The $\Omega$ has the same conditional
posterior distribution as the model parameter $H$ of
\citet{marttinen2014assessing}.

\subsection*{Improved update of $\Psi$}
The computational bottleneck of the na{\"i}ve Gibbs sampler is the update of
parameter $\Psi$, which has $P S_1$ elements with a joint multivariate Gaussian distribution, conditionally on the 
other parameters \citep{Geweke96,marttinen2014assessing}. Thus, the inversion of the precision matrix of the joint distribution
has a computational cost of $O(P^3S_1^3)$. To remove the bottleneck, we reparameterize our model, after
which a linear algebra trick by \citet{stegle2011efficient} can be used to  reduce the computational cost of
the bottleneck to $O(P^3\ + S_1^3)$. When sampling $\Psi$ we also integrate
over the distribution of $\Omega$ following the standard result from Equation
(\ref{eq:integrate_over_Omega}). The reparameterization and the new posteriors are presented in the
Supplementary material. 

In brief, the trick is that the eigenvalue decomposition of a matrix of the form 
\begin{equation}
  C \otimes R + \sigma I \label{eq:stegle_trick}
\end{equation}
can be evaluated inexpensively. After reparameterizing the model in the proposed way the posterior covariance matrix of $\Psi$ becomes of the
form (\ref{eq:stegle_trick}) and the eigenvalue decomposition can then be used to efficiently generate
samples from the posterior distribution of $\Psi$. We note that the trick can also be applied to the
original formulation of the Bayesian reduced-rank regression model by \citet{Geweke96} and the R-code
published with this article allows generating samples from the original model as well. In the next Section, we
compare the computational cost of the algorithm using the na{\"i}ve Gibbs sampler and the improved version
that uses the new parameterization.
\subsection*{Sampling the maximum rank of the model}
The sparse infinite factor analysis model presented by \citet{bhattacharya2011sparse} uses a certain adaption procedure to update the maximum rank, i.e., the truncation point of their infinite-rank factor model. The idea is to update the maximum rank occasionally during the algorithm such that ranks having all elements of the corresponding projection vectors within some pre-specified distance from zero are removed from the model and, if none of the ranks has all elements within the threshold, another rank is added into the model. We have implemented a modification of this approach where we adapt the maximum rank of our infinite reduced rank regression model using a pre-specified cutoff for the amount of variance explained by the corresponding rank. With a slight abuse of terminology, we shall call this updating of the rank as ‘sampling’ in the sequel.

\section{Experiments} \label{sec:realdata}

We start with a basic validation of the latent-noise BRRR model, and its relative merits over alternatives in a 
prediction task, using simulations with the ground truth available (Section \ref{sec:simulation_exp}), and a 
real-world omics dataset (Section \ref{sec:NFBC_res}). Section \ref{sec:NFBC_detailed_res} analyses these results 
in more detail and identifies the characteristics of the proposed latent-noise BRRR model that are responsible for 
the performance differences observed, by considering the impact of each novel model aspect in isolation. 
Section (\ref{sec:association_exp_res}) investigates another application domain, the detection of multivariate 
associations. In order to assess the prediction performance in more general, we analyse several additional 
real-world data sets from different domains in Section \ref{sec:other_data_sets}.

Different aspects of the inference algorithm are considered in three sub-sections: sampling vs. cross-validation of 
the rank and the latent signal-to-noise ratio (Section \ref{sec:inf_proc_eval}), speedup resulting from the 
proposed re-parameterization of the algorithm (Section \ref{sec:runtime_experiment}), and convergence 
diagnostics (Section \ref{sec:efficiency_of_the_algorithm}). To assess the value of further extensions, 
Section \ref{sec:latent+independent_BRRR} considers a model that includes both latent and independent structured 
noise simultaneously. Finally, Section \ref{sec:overview} summarizes the findings on all real data sets.

\subsection{Data sets} \label{sec:data_sets}
Experiments were performed on the following data sets:
\begin{itemize}
 \item[] \textbf{NFBC1966} $[N = 4702, P = 101, K = 96, \text{metabolomics prediction from SNPs}]$ The NFBC1966 
data set comprises genome-wide SNP data along with metabolomics measurements for a cohort of 4,702 
individuals \citep{rantakallio1969groups,soininen2009high}. With these data, 96 metabolites belonging to the 
subclasses VLDL, IDL, LDL and HDL \citep{inouye2012novel} were used as the target variables and SNPs known to 
be associated with lipid  metabolism \citep{teslovich2010biological,kettunen2012genome,Genetics_Consortium13} 
were used as the covariates. Effects of age, sex, and lipid lowering medication were regressed out from the 
metabolomics data as a preprocessing step. For the genotype data, SNPs with low minor allele frequency 
($<$0.01) were removed as a preprocessing step. For this data set, the comparison method GFlasso 
required excessive training time and we used 5-fold cross-validation to evaluate test set performances. Where 
cross-validation was needed for selecting model parameter values, the  validation data performance was measured as an 
average over 3 validation sets, each comprising $\frac{1}{10}$ of the training samples.
\item[] \textbf{DILGOM} $[N = 509, P = 65, K = 18 \ldots 137, $ metabolomics and gene expression prediction 
from SNPs$]$ 
The DILGOM data set \citep{inouye2010metabonomic} consists of genome-wide SNP data along with metabolomics and
gene expression measurements. For details concerning metabolomics and gene expression data collection, see
\citet{soininen2009high} and \citet{kettunen2012genome}. In total 509 individuals had all three measurement
types. The DILGOM metabolomics data comprises 137 metabolites, most of which represent NMR-quantified levels
of lipoproteins classified into 4 subclasses (VLDL, IDL, LDL, HDL), together with quantified levels of amino
acids, some serum extracts, and a set of quantities derived as ratios of the aformentioned metabolites. All 
137 metabolites were used simultaneously as prediction targets. In gene expression prediction, in total 387 
probes corresponding to curated gene sets of 8 KEGG lipid metabolism pathways were used as the prediction 
targets. A separate model was learnt for each pathway. The average number of probes in a pathway was 48. For 
details about the pathways, see the Supplementary material. On these data sets, 10-fold cross-validation was 
used to evaluate test set performances. To select values of the parameters that required evaluation on 
validation data, the training data was then further divided into 9 folds, on which cross-validation was 
performed to select parameters according to averaged validation set performance.

\item[] \textbf{fMRI} $[N = 1307, P = 776, K = 250, \text{fMRI response prediction from text stimuli}]$ 
The cognitive neuroscience data set \citep{10.1371/journal.pone.0112575} consists of a 
time series of fMRI measurements from 8 subjects reading a chapter from ``Harry Potter and the Sorcerers 
Stone``  using \emph{Rapid Serial Visual Presentation}: words of the text are presented one by one in the 
center of a screen. Brain voxel activations were measured every 2 seconds. The 250 most accurately predictable 
voxels \citep[see Supplementary material of][]{10.1371/journal.pone.0112575} of the fMRI measurements were used as 
prediction targets. The fMRI measurements from all 
patients were predicted simultaneously from features of the words being shown, such as semantic and syntactic 
properties, visual properties and discourse level features. The data were divided into 10 folds, only two of 
which were used to measure test data performance. This computational compromise was needed as the 
preprocessing \citep{10.1371/journal.pone.0112575} for each fold required about 10,000 hours of 
computation. To select the values of parameters that required evaluation on validation data, the training data 
were further divided into 10 folds, on which cross-validation was performed to select parameters according 
to averaged validation set performance.

\item[] \textbf{econ} $[N = 120, P = 52, K = 52, \text{macroeconomic time series prediction}]$ The macroeconomic 
time series data set \citep{stock2006forecasting} consists of 
monthly values of 52 macroeconomic indicators. Prediction performance of these values from their earlier values was measured with different lags (1 month, 2 months, etc.). The data were processed 
as described by \cite{carriero2011forecasting}. Data for each month were used as a test set (395 test sets) 
while using data from the previous 10 years for training. Where cross-validation was needed for 
learning the values of model parameters, data from the last 2 years before the month-to-be-predicted were 
used for validation and data from the previous 8 years for training.

\end{itemize}

\subsection{Methods included in comparison} \label{sec:methods_in_comp}
We compared the latent-noise BRRR with a state-of-the-art sparse multiple-output regression method Graph-guided 
Fused Lasso ('GFlasso')  \citep{Kim09}, BRRR/factor regression model \citep{marttinen2014assessing} with and 
without the \emph{a priori} independent noise model ('independent-noise BRRR', 'BRRR without noise model'), 
standard Bayesian linear model ('blm') \citep{gelman2004bayesian}, elastic-net-penalized multi-task learning 
('L2/L1 MTL'), kernel regression with linear and Gaussian kernels combined with a process for removing 
confounding factors \citep{stegle2012using} ('KRR with linear kernel + PEER', 'KRR with Gaussian kernel + 
PEER') and a baseline method of predicting with target data mean. GFlasso constitutes a suitable comparison as 
it encourages sharing of information between correlated responses, as our model, but does that within the 
Lasso-type penalized regression framework without the use of a noise model to explain away the structured 
noise. L2/L1 MTL is a multitask regression method  implemented in the \texttt{glmnet} ~package \citep{glmnet} 
that allows elastic net regularization. It does not use a noise model to explain away
confounders either. The blm was selected as a simple single-task baseline.

In one of the experiments, on an association study, latent-noise BRRR is compared with independent-noise BRRR and canonical 
correlation analysis ('cca'), considered the state-of-the-art methods for the detection of 
multivariate associations \citep{Marttinen2013,marttinen2014assessing}. Additionally, the simple univariate linear 
model ('lm') is included as it represents the common baseline in association analysis.

We compare latent-noise BRRR also with two other new models for structured noise modeling. In the simulations, we study 
the performance of correlated Bayesian reduced rank regression ('correlated BRRR'), which is presented in more detail 
in the Supplementary material. In brief, in the correlated BRRR, the correlation structure of the target variables learnt 
by an \emph{a priori} independent noise model is used as a prior for the regression weight parameters. With the 
NFBC1966 data and the macroeconomic time series data sets, we also study the performance of the method presented 
in Equation (\ref{eq:latent+independent_noise_model}) in Section \ref{sec:latent_independent_differences} that 
explicitly models both latent and independent structured noise, abbreviated as 'latent+independent-noise BRRR'.

Parameters for the different methods were specified as follows:
\begin{itemize}
 \item[] \textbf{GFlasso:} The regularization parameters of the \texttt{gw2} model were selected from the
default grid using cross-validation. The method has been developed for genomic data indicating the
default values should be appropriate. However, for NFBC1966 data, we were unable to run the method with the
smallest values of the regularization parameters $\{110, 60, 10\}$ due to lengthy runtime with these values. 
With this computational compromise of leaving out these three values, the average training time for the largest training data sets was $\sim$ 
650 h. With NFBC1966 data, the pre-specified correlation network required by the GFlasso was constructed to match 
the VLDL, IDL, LDL, and HDL metabolite clusters from \citet{inouye2012novel}. Within these clusters, the 
correlation network was fixed to the empirical correlations, and to 0 otherwise. With DILGOM data, we used 
the empirical correlation network, with correlations below 0.8 fixed to 0 to reduce the number of edges in 
the network for computational speedup.
\vspace{1.5mm}

 \item[] \textbf{independent-noise BRRR, BRRR without noise model:} Hyperparameters $a_1$ and $a_2$ of all 
the BRRR models were fixed to 10 and 4, respectively. In total 1,000 MCMC samples were generated and 500 were 
discarded as burn-in. In preliminary tests similiar results were obtained with 50,000 samples. The remaining 
samples, thinned by a factor of 10, were used for prediction. The maximum rank of the infinite-rank BRRR 
model was learned using cross-validation from the set of values $\{5,10,15\}$ for the NFBC1966 data set, 
$\{2,4,8\}$ for the metabolomics prediction task on the DILGOM data set and $\{2,5,10,20\}$ for the gene 
expression prediction task on the DILGOM data set. These grids were selected based on initial experiments. 
For the fMRI response prediction, the possible values for the maximal rank were limited to $\{2,4\}$ in 
order to save computational time. For the econometrics data set, maximum ranks of $\{5,10,20\}$ were used. 
In the associaton detection task, the rank of independent 
noise BRRR was fixed to 1 as this was already sufficient for the task.
\vspace{1.5mm}

 \item[] \textbf{latent-noise BRRR:} With the NFBC1966 data, the latent signal-to-noise ratio $\beta$ was
selected using cross-validation from a range of values from $100$ to $\frac{1}{100}$,  
$\beta = \{100,10,2,1,\frac{1}{7.5},\frac{1}{15},\frac{1}{30},\frac{1}{60},\frac{1}{100}\}$, in order to 
thoroughly evaluate the sensitivity of the model to this parameter. For the other data sets/tasks, the sets of 
values were as follows: DILGOM metabolomics prediction: $\beta=\{10,2,1,\frac{1}{7.5},\frac{1}{15}, 
\frac{1}{30},\frac{1}{60},\frac{1}{100}\}$, DILGOM gene expression prediction 
$\beta=\{10,1,\frac{1}{5},\frac{1}{10},\frac{1}{30},\frac{1}{50},\frac{1}{100},\frac{1}{300}\}$ and for 
macroeconomic time series prediction 
$\beta=\{10,2,1,\frac{1}{7.5},\frac{1}{15},\frac{1}{30},\frac{1}{60},\frac{1}{100}\}$. For fMRI response 
prediction, the set of values was limited to $\beta=\{10,1,\frac{1}{10}\}$ to save computation time. Other 
parameters, including the number of iterations, were set as for the independent-noise BRRR. The performance of 
the model was evaluated both by sampling the maximum rank and 
by learning it with cross-validation from the same range of values as 
with independent-noise BRRR. Shrinkage hyperparameters were set to noninformative values, $a_1 = $10 and $a_2 = 
$ 4, similarly to the corresponding parameters $a_3$ and $a_4$ of independent-noise BRRR.
\vspace{1.5mm}

 \item[] \textbf{blm:} The variance hyperparameter of BLM was integrated over using MCMC. The 
variance hyperparameter was assigned a Gamma prior with both shape and rate parameters set to 1. 
In total 1,000 posterior samples were generated and 500 were discarded as burn-in.
 \vspace{1.5mm}

 \item[] \textbf{L1/L2 MTL:} The effects of different types of regularization penalties are an active research
topic and we ran a continuum of mixtures of L1 and L2 penalties ranging from group lasso to 
ridge regression. The mixture parameter $\alpha$ controlling the balance between L1 and L2 regularization was
evaluated on the grid [0, 0.1, $\ldots$, 0.9, 1.0] and selected using a 10-fold cross validation. 
The default convergence threshold parameters of \texttt{glmnet} were used and no warnings/numerical problems 
occured.

 \item[] \textbf{KRR with linear kernel + PEER:} First, the PEER software \citep{stegle2012using} was used to remove 
the effects of confounders using 15 components. Then kernel ridge regression with a normalized linear kernel 
\citep{bishop2006pattern} was applied using the residuals from PEER as the target variables. Kernel ridge 
regression was regularized according to the standard approach of adding parameter $\lambda$ to the diagonal 
elements of the kernel. The value of $\lambda$ was selected using cross-validation from a set of 10 values 
ranging from 0.1 to 100, $[10^{-1}, 10^{-0.66}, \ldots, 10^{1.67}, 10^{2}]$. To share 
information between the different target variables, the approach of using the same kernel for all 
target variables was adopted.

 \item[] \textbf{KRR with Gaussian kernel + PEER:} Kernel ridge regression using a Gaussian kernel 
was used. Regularization and the use of PEER were otherwise similar to KRR with linear kernel + 
PEER. The radius parameter of the Gaussian kernel was selected using cross-validation from a set of 30 
values ranging from 0.001 to 1000, $[10^{-3}, 10^{-2.79}, \ldots, 10^{2.79}, 10^{3}]$

\item[] \textbf{cca:} This is the conventional classical canonical correlation analysis that attempts to identify 
linear combinations of the columns of the input and output matrices that are maximally correlated with each other. 

\item[] \textbf{correlated BRRR:} Rank and hyperparameters $a_1$, $a_2$, $a_3$ and $a_4$ were set as with the 
independent-noise BRRR. This model is presented in detail in Supplementary Section 1.

\item[] \textbf{latent+independent-noise BRRR:} With the NFBC1966 data, the hyperparameters $a_1$, $a_2$, $a_3$ and 
$a_4$ were set as with the independent-noise BRRR. The latent signal-to-noise ratio $\beta$ was
selected using cross-validation from a range of values from $100$ to $\frac{1}{100}$,  
$\beta = \{100,10,2,1,\frac{1}{7.5},\frac{1}{30},\frac{1}{60},\frac{1}{100}\}$ and the maximum rank was fixed to 10.
For the econometrics data set, maximum ranks of $\{5,10,20\}$ were used and  the signal-to-noise ratio $\beta$ was
selected using cross-validation from the values $\beta = \{10,2,1,\frac{1}{5},\frac{1}{10},\frac{1}{30}$. For both 
data sets, the variance parameter of the \emph{a priori} independent noise $H$ was selected from values 
$\{10^{-6},1\}$, value $10^{-6}$ corresponding to the extreme case of latent-noise BRRR.

\end{itemize}

\subsection{Simulation experiment: impact of the noise model assumptions} \label{sec:simulation_exp}
In this Section, we study the implications of different noise model assumptions. Performances of models with 
different noise model assumptions are measured on simulated data sets generated from a continuum of models 
between the two extremes of assuming either latent noise, or \emph{a priori} independent regression and noise 
models. The synthetic data are generated according to 
\begin{equation}
 Y = (X \Psi + \alpha \Omega) \; \Gamma  + (1-\alpha) H \Lambda + E,
\label{eq:mixture_model}
\end{equation} 
where $\text{vec}(E) \sim \mathcal{N}(0,I_{NK})$ and the parameter $\alpha \in [0,1]$ defines the proportion of variance 
attributed to the latent noise versus independent noise. We study a continuum of problems with the values of parameter $\alpha = 0, 
0.1, \ldots, 1$.
The parameters $\Gamma$ and $\Lambda$ are orthogonalized using Gramm-Schmidt orthogonalization. The 
parameters 
are scaled so that covariates $X$ explain 3 \% of the variance of $Y$ through $ X \Psi \Gamma$, the diagonal
Gaussian noise $\mathcal{N}(0,I_{NK})$ explains 20 \% of the total variance of $Y$ and the structured noise
$\alpha \Omega \Gamma  + (1-\alpha) H \Lambda$ explains the remaining 77 \% of the total variance of $Y$. The
simulation was repeated 100 times and training data sets of 500 and 2000 samples were generated for each
replicate. To compare the methods, performance in mean squared error (MSE) of the models learned with each
method was compared to that of the true model on a test set of 15 000 samples. The number of covariates was 
fixed to 30 and the number of dependent variables to 60. Rank of the regression coefficient matrix and 
structured noise was set to 3 when simulating the data sets.

For independent-noise BRRR, the rank of the regression coefficient parameters $\Psi$ and $\Gamma$ was fixed 
to the true value while the rank of the noise model was learnt from the data. For latent-noise BRRR, the 
performance of the model was evaluated both by fixing the rank of the regression coefficient matrix to its 
true value and by learning it from the data. The variance of $\Omega$ was
selected using 10-fold cross-validation. The grid for latent signal-to-noise ratios $\beta$ was $\beta = \frac{1}{5}$ to $\frac{1}{15}$, $\beta = \{\frac{1}{5}, 
\frac{1}{7.5}, \frac{1}{10}, \frac{1}{12.5}, \frac{1}{15} \} $. More specifically, $\text{Var}(\text{vec}(\Omega)) = 
\sigma_{\Omega}^2 I_{NK}$ where $\sigma_{\Omega}^2 = \frac{1}{\beta} \times \text{Trace}\left(\text{Var}\left( 
X \right)\right)$. The grid was chosen according to the 
interpretation given in Section \ref{sec:selecting_var}; it corresponds to assuming that the latent noise 
explains 5 to 15 times the variance explained by the covariates.

\begin{figure}[ht!]
\centering
\begin{tabular}{cc}
. \includegraphics[width=0.5\textwidth]{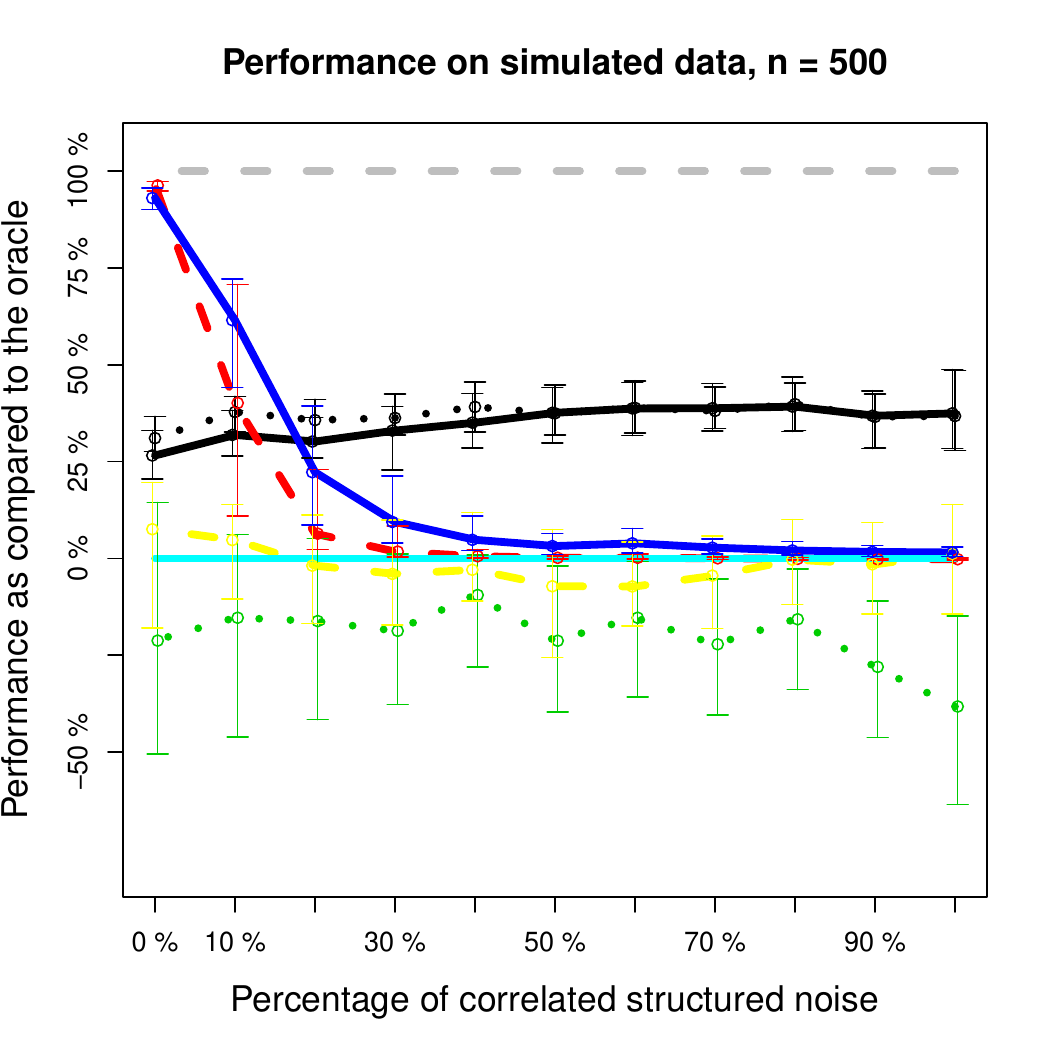} 
&
 \includegraphics[width=0.5\textwidth]{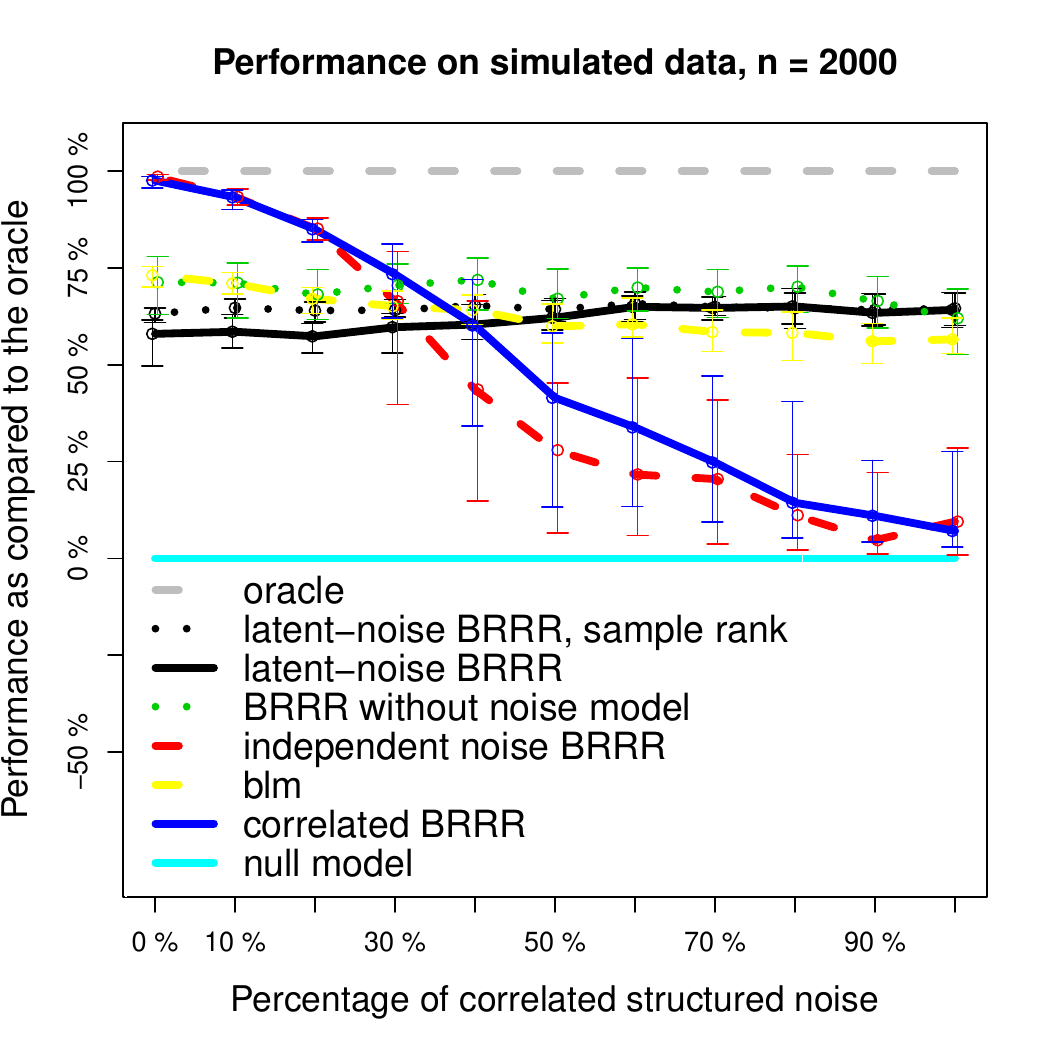} \\
(a) & (b) \\
\end{tabular}
\caption{Performance of different methods, compared to the true model, as a function of the proportion of
latent noise with a training set of (a) 500 and (b) 2000 samples. The x-axis indicates the proportion of noise
generated according to the latent noise assumptions (100\% corresponds to $\alpha = 1$). Bars denote $\pm$ 1 
standard deviation, computed independently for each x-coordinate. The performance of 100\% means the amount of variance explained by the model is equal to the amount explained by the true model. The performance of 0\% means that the method does not explain any variance of the target variables, whereas negative values indicate the variance actually increases after taking the predictions into account.}

\label{Fig_simulaatio}
\end{figure}

Figures \ref{Fig_simulaatio} (a) and (b) present the results of a simulation study with training sets of 500 
and 2000 samples, respectively. When the structured noise is generated according to the conventional 
assumption of independent signal and noise, the model making the independece assumption (i.e., the independent-noise BRRR) performs equally well to
the true model with both 500 and 2000 samples. However, when the assumption is violated and the proportion of
latent noise increases, the performance of the independent-noise BRRR
breaks down, whereas the latent-noise BRRR performs consistently well.  The method that does not explain 
away the structured noise at all (BRRR without noise model)  is always inferior to the null model with the training set of 500 samples. When 
the number of training samples is increased to 2000 and the noise is generated according to the latent-noise 
assumption, the model, however, outperforms even the independent-noise BRRR.Thus, having no noise model is in this 
case better than having the noise model based on the incorrect independence assumption, which emphasizes the 
importance of the assumptions on which the noise model is based. Interestingly, with n=2000 the BRRR without 
noise model is among the best performing methods whereas with n=500 it is clearly the worst, highlighting the fact 
that the smaller n gets, the more important the right assumptions become. 

The latent-noise end of the continuum appears to be more difficult for the methods that do not account for the 
structured noise (blm, BRRR without noise model). This weak but consistent trend can be seen in 
Figure \ref{Fig_simulaatio}(b) where the difference between the oracle and these methods increases with the 
percentage of latent noise. This behaviour is, however, rather intuitive in terms 
of Equation (\ref{eq:mixture_model}); by rewriting
\begin{align*}
 Y &= (X \Psi + \alpha \Omega) \; \Gamma  + (1-\alpha) H \Lambda + E, \\
   &= X \Psi \Gamma + \alpha \Omega \Gamma  + (1-\alpha) H \Lambda + E \\
\end{align*}
it is obvious that as $\alpha \rightarrow 1$, the structured noise (coming from $\Omega$ and $H$) will with 
certainty be projected on the particular target variables that are affected by the covariates $X$. In other words, 
latent noise blurs exactly the relationships of interest, being very disruptive. 

Figure \ref{Fig_simulaatio} shows results also for an alternative novel model that shares information between 
the noise and regression models (correlated BRRR, see Supplementary material for a detailed description). The 
model includes a separate noise model for the structured noise, as in (\ref{eq:factor_regression_modeling}), 
but achieves the information sharing by assuming a joint prior for the noise and regression models. In 
detail, conditional on the noise model, the current residual correlation matrix between the response variables 
is used as a prior for the rows of $\Gamma$. This way the correlations between target variables are propagated 
into the corresponding regression weights; however, the strongest noise components are not automatically 
coupled with the strongest signal components. Notably, the performance of the correlated BRRR model is very 
similar to the regular BRRR model that does not have any dependence between the noise and signal components. 

\subsection{NFBC1966: metabolomics prediction} \label{sec:NFBC_res}
\begin{figure}[ht!]
\centering

 \includegraphics[width=0.8\textwidth]{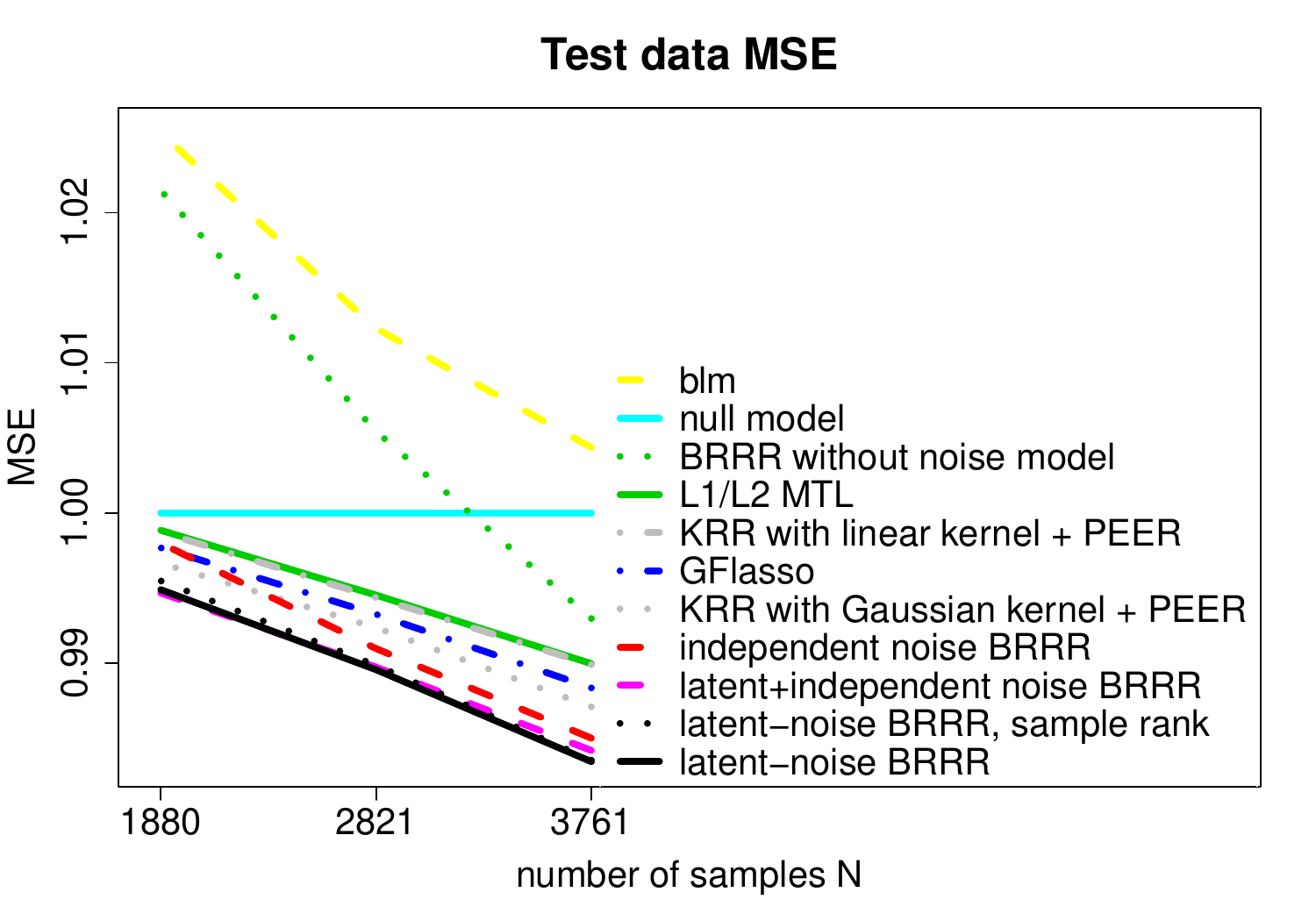}

\caption{Test data MSE for different amounts of training data on the NFBC1966 metabolomics data.}
\label{Fig_NFBC66_tulokset}
\end{figure}

In this Section, the models accounting for latent noise are evaluated in terms of predictive 
performance on the NFBC1966 data with different training set sizes. Figure \ref{Fig_NFBC66_tulokset} 
presents the test data MSE for the different methods. With all training set sizes, latent-noise BRRR 
outperforms the other methods. Method blm performs worse than the baseline (null model), even with the 
largest training data set containing 3761 individuals, and BRRR without noise model requires the largest training 
set size in order to outperform the baseline. A paired t-test for the 
performance difference between latent-noise BRRR and independent-noise BRRR yields a p-value of 0.03 
suggesting a statistically significant difference.

\subsection{Differences between latent-noise BRRR and independent-noise BRRR on NFBC1966 metabolomics 
prediction} \label{sec:NFBC_detailed_res}
The two differences between our new approach, latent-noise BRRR, and independent-noise BRRR are 
(1) model structure (latent-noise BRRR shares parameters between the regression and noise models) and (2) using
the latent signal-to-noise ratio parameter $\beta$ to regularize the model. In order to identify how these 
developments lead to the observed performance differences on the NFBC1966 data, we performed a sensitivity 
analysis for the two methods with respect to the assumed amount of variance attributed to the noise model.

\begin{figure}[ht!]
\centering
\begin{tabular}{cc}
 \includegraphics[width=0.5\textwidth]{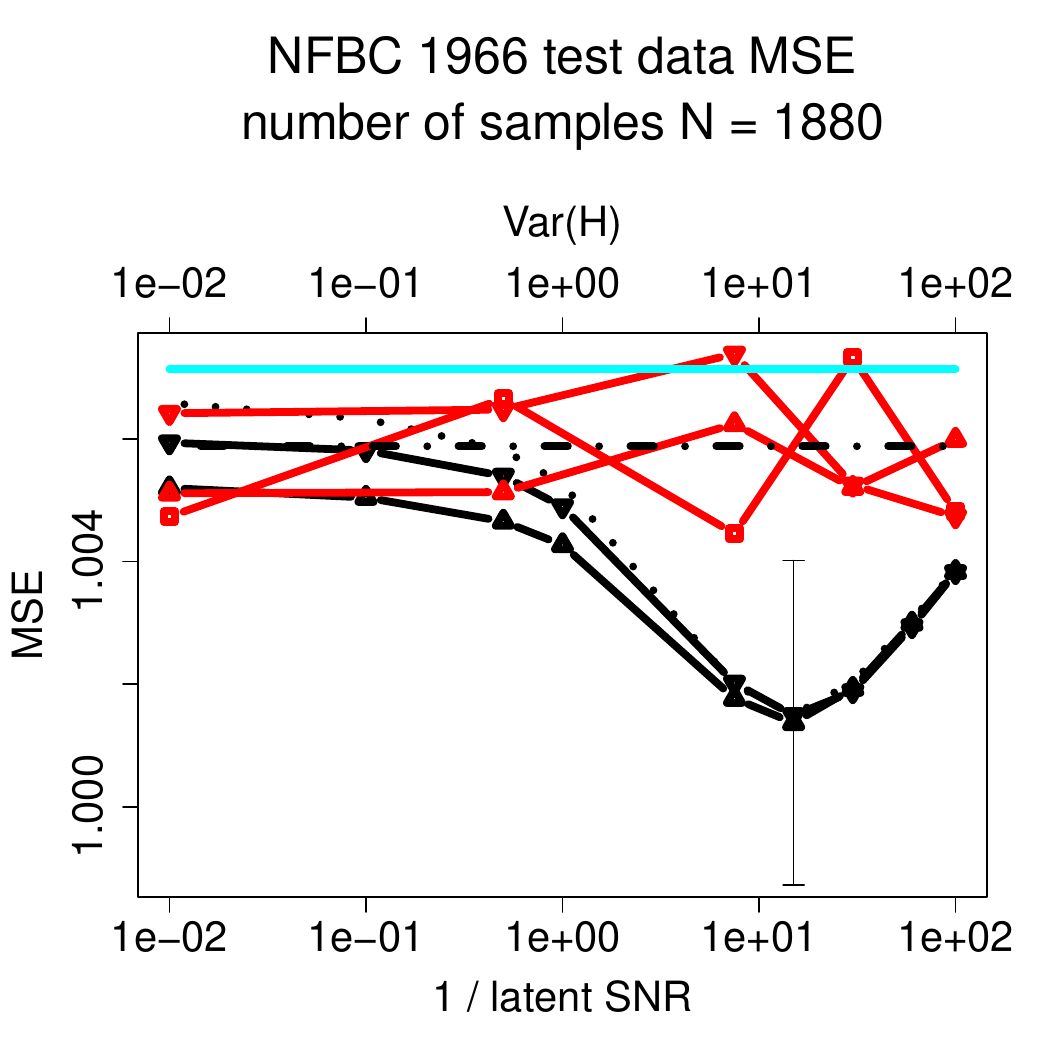} &
 \includegraphics[width=0.5\textwidth]{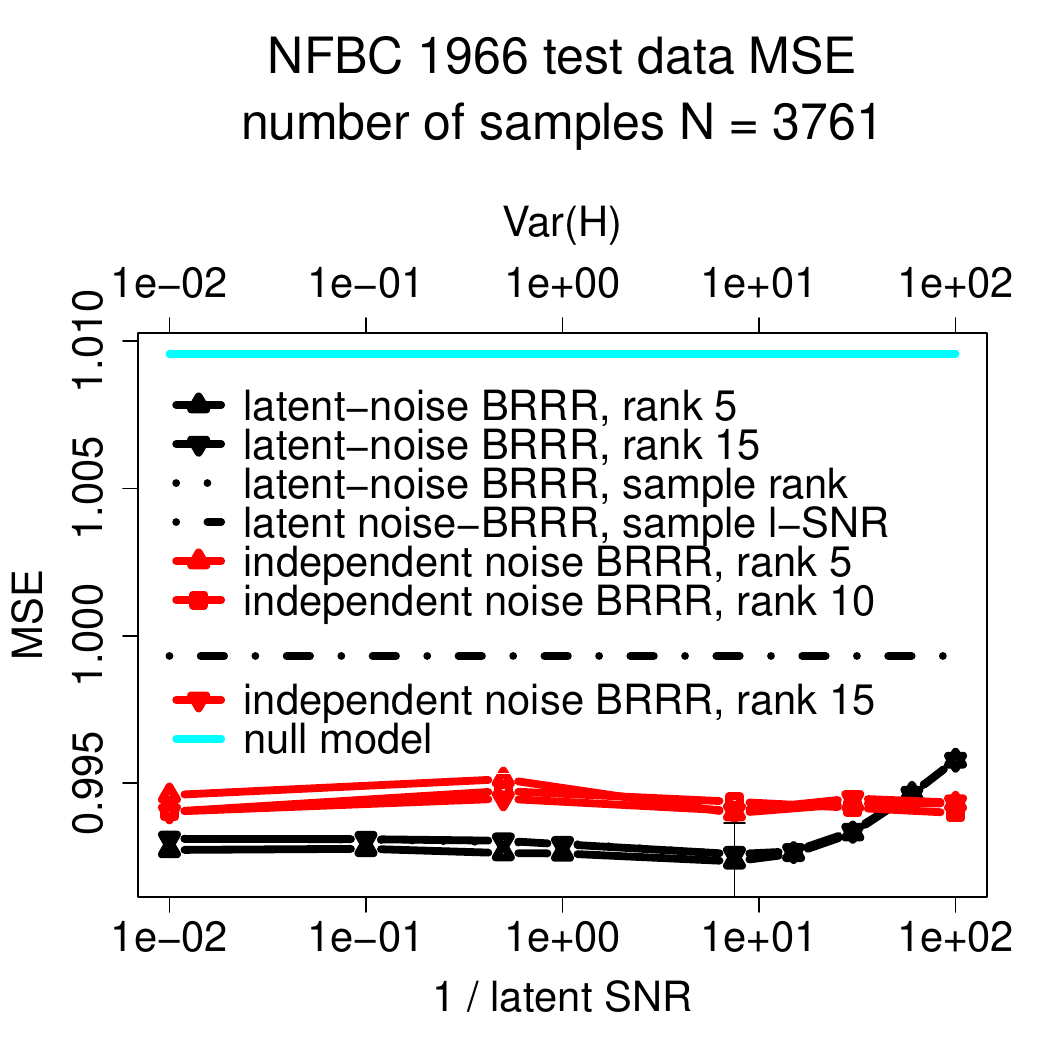} \\
\end{tabular}
\caption{Sensitivity of latent-noise BRRR and independent-noise BRRR to the variance of the structured noise 
with different maximum ranks. The results are on NFBC1966 test data MSE ($N = $1880 and $N = $3761) 
as a function of the noise model variance. Lower axis: \emph{a priori} latent signal-to-noise ratio of 
latent-noise BRRR and the upper axis: variance of the model parameter $H$ of independent-noise BRRR. The bar 
denotes the standard deviation of the test set performance difference observed between the two models in 
cross-validation. The figures also present the performance of the null model and the performance of the
latent-noise BRRR when using sampling to infer the latent signal-to-noise ratio (latent-noise BRRR, sample l-SNR)
and when using sampling to infer to infer the maximum rank (latent-noise BRRR, sample rank). When $N = 3761$, sampling 
the rank results in similar performance as obtained with the fixed values and thus the curves overlap.}
\label{Fig_NFBC66_vertailu_tulokset}
\end{figure}

Figure \ref{Fig_NFBC66_vertailu_tulokset} presents the results of this sensitivity analysis. For latent-noise BRRR, 
the assumed variance of the noise model controlled by the \emph{a priori} signal-to-noise ratio $\beta$ affects performance 
in a consistent way, whereas for independent-noise BRRR the impact appears random. If the 
performance difference stemmed mainly from controlling the variance of the noise model, controlling that
parameter for both models should lead to similar results. On the other hand, if the difference in the model structure alone sufficed to explain the performance difference, the difference should not be sensitive to the variance of the noise model. Hence, we conclude 
that, on this data set, both the new model structure and regularization by using the latent signal-to-noise ratio are 
required for improved performance.

\subsection{Evaluation of the chosen inference procedures for rank and noise parameters} \label{sec:inf_proc_eval}
Inference for the proposed model could naturally be done in several alternative ways. In this Section we justify 
the proposed inference procedure. 

In the simulations (Section \ref{sec:simulation_exp}), sampling the maximum rank of the infinite prior worked 
well, measured in terms of predictive performance. Figure \ref{Fig_simulaatio} shows that sampling 
the maximum rank actually improves performance, as compared to fixing it to the value used in the generative 
process,  when the latent noise assumption is wrong (left end), both when $N =$ 500 and when $N 
=$ 2000. When the latent noise assumption holds (right end), the two inference 
procedures perform equally well. With the NFBC1966 data set (Figure \ref{Fig_NFBC66_tulokset}), learning the 
maximum rank of the infinite prior by sampling (latent-noise BRRR, sample rank) or by 
cross-validation (latent-noise BRRR) results in very similar test set performances, similarly to in the simulation 
experiment in Section \ref{sec:simulation_exp}. Hence, we conclude that for learning the maximum rank, both sampling and cross-validation are appropriate techniques. We also ran the independent-noise BRRR so that the rank was 
sampled instead of selecting it using cross-validation on this data. However, 
the results were poor, with the test data MSE  equal to 1.019 with $N = $1880 and 1.005 with $N = $3761. 
The lines were omitted for clarity. We hypothesize that the problems with the instability of the model (see Section \ref{sec:efficiency_of_the_algorithm}) were accentuated when the rank was sampled.

The key parameter of our model, the latent signal-to-noise ratio, was estimated using cross-validation. 
In the simulations, the cross-validation based scheme allowed estimation of the latent signal-to-noise ratio to a 
reasonable accuracy. The estimated values are included in Figure 2 in the Supplementary material. While 
the latent signal-to-noise ratio of the generative process was $\approx$ 1/25, the estimated posterior latent 
signal-to-noise ratios ranged from $\frac{1}{14}$ to $\frac{1}{19}$ in the parts of the domain where 
the percentage of correlated structured noise was 100-80\%. When the percentage of correlated structured noise was 
0-10 \%, the model correctly learnt lower variance for the latent noise and a corresponding stronger latent 
signal-to-noise ratio $\beta$.
 
We also studied the performance of latent-noise BRRR while sampling the variance of the noise model. A 
noninformative prior was assigned for the variance of $\Omega$, $\Omega \sim \mathcal{N}(0, \sigma_{\Omega}^2)$ 
and $ \sigma_{\Omega}^{-2} \sim \text{Gamma}(\text{shape = 0.001, rate = 0.001})$. The performance of this 
model is presented in Figure \ref{Fig_NFBC66_vertailu_tulokset}. The 
performance of latent-noise BRRR when samping the variance of $\Omega$ is consistently worse than when using 
cross-validation to select the value of the latent signal-to-noise ratio. Hence, we conclude that, as opposed to other parameters, cross-validation is needed to learn the latent-signal to ratio to reach the improved performance.

\subsection{NFBC1966: multivariate association detection} \label{sec:association_exp_res}
Detection of associations between multiple SNPs and metabolites is a topic that has received attention 
recently \citep[see, e.g.,][]{kim2009multivariate,inouye2012novel,marttinen2014assessing}. Here 
we demonstrate the potential of the new method in this task using two illustrative example genes for which 
ground truth is available. Associations between SNPs within two genes, \emph{LIPC} and \emph{XRCC4}, and 
the metabolites in the NFBC1966 data are investigated in the experiment. Note that the covariates (SNPs) used in this 
experiment are different from the ones used in the prediction experiment: here SNPs in individual genes are used, whereas 
in the prediction experiment all known lipid-associated SNPs were used. \emph{LIPC} was selected as a 
reference, because it is one of the most strongly lipid-associated genes. On the contrary, \emph{XRCC4} was 
discovered only recently using three cohorts of individuals \citep{marttinen2014assessing}, and it was 
selected to serve as an example of a complex association detectable only by associating multiple SNPs with 
multiple metabolites, and not visible using simpler methods.

We use the proportion of total variance explained (PTVE) as the test score \citep{marttinen2014assessing}, and
sample 100 permutations to measure the power to detect the associations. Furthermore, we use downsampling to
evaluate the impact of the amount of training data. For comparison, we select the BRRR, the exhaustive 
pairwise (univariate) linear regression ('lm'), and canonical correlation analysis (CCA) 
\citep{ferreira2009multivariate}, these being the methods that have been proposed for the task and having a 
sensible runtime in putative genome-wide applications. For lm, the minimum p-value of the regression 
coefficient over all SNP-metabolite pairs, and for the CCA, the minimum p-value over all SNPs (each SNP 
associated with all metabolites jointly) are used as the test scores. The association involving the 
\emph{XRCC4} gene was originally detected using the BRRR model;  however, unlike here, informative priors were 
used for the regression coefficients.

Table \ref{Table_permutaatiot} presents the ranking of the original data among the permuted data with
different sample sizes and methods. Ten MCMC chains were computed for both models to account for sampling variability 
on this difficult and relatively strongly collinear data. The association score was obtained by averaging over the scores for different chains. As expected, all methods were able to detect the association 
involving  \emph{LIPC} with both training set sizes. However, latent-noise BRRR had the highest power 
to detect the \emph{XRCC4} gene. 

\vskip 0.15in
\begin{table}
 \centering
\small
\caption{Power of different methods to detect the association between metabolomics profiles and \emph{XRCC4} or \emph{LIPC} genes with $N
= $4702 and $N = $2351 samples. Power is measured as the
proportion of association test scores in permuted data sets smaller than the test score in the original data
set. Value 1 indicates that the association score of the unpermuted data was higher than the score in any
permutation.}
\label{Table_permutaatiot}
\vskip 0.15in
\begin{tabular}{l|cc|lcc}
 & \emph{XRCC4} &  & \emph{LIPC} & \\
 & $N = $4702 & $N = $2351 &  $N = $4702 & $N = $2351  \\
\hline
latent-noise BRRR & 0.98 & 0.94 & 1 &  1.00 \\
independent-noise BRRR & 0.41 & 0.32 & 1 &  0.99 \\
lm & 0.62 & 0.74 & 1 &  1.00 \\
cca & 0.20 & 0.24 & 1 & 1.00 \\
\hline 
\end{tabular}
\vskip -0.1in
\end{table}

\subsection{Results: other real-world data sets} \label{sec:other_data_sets}
To thoroughly study the empirical value of the new method, we compared it to alternative methods on macroeconomic time 
series prediction, metabolomics and gene expression prediction experiments on the DILGOM data set and the fMRI 
response prediction. In these domains, explaining away structured noise is of crucial importance.

With the DILGOM data, the prediction of the weak effects was challenging for all methods. Indeed, we
noticed that the null model using the average training data value for prediction was better than any other method 
in terms of MSE over all target variables with the single exception of L1/L2 MTL, which set all regression 
coefficients to zero thus reducing to the null model. However, a detailed investigation of the results revealed that 
while many of the target variables could not be predicted at all (as indicated by the worse than null model MSE) 
some of the target variables could still be predicted better than the null model, and by focusing the analysis 
on the MSE computed over the predictable target variables (\textit{i.e.,} those that could be predicted
better than the null by at least one method), comparisons regarding the model performances could still be 
made. For consistency, both metrics were computed also with the fMRI and econometrics data sets. To save computation 
time, we chose to evaluate only the cross-validation based variant of our model for the fMRI data, as this approach 
had already been identified as the most promising implementation of our method.

Table \ref{table:other_data_sets_MSE_predictable} and supplementary Table 1 present 
the results of the macroeconomic time series prediction experiment, metabolomics and gene expression
prediction experiments on the DILGOM data set and the fMRI response prediction experiments. The results have been 
normalized so that the score for the null model (prediction using the mean) is 1. Table 
\ref{table:other_data_sets_MSE_predictable} presents the results for the predictable target variables and 
supplementary Table 1 presents
the results obtained by averaging test data MSE over all target variables. 

Latent-noise BRRR outperforms independent-noise BRRR consistently on the gene expression (on 8/10 folds), metabolomics 
(10/10 folds) and fMRI response prediction (2/2 folds) tasks on both scores. In the fMRI response prediction, 
the latent-noise BRRR and L1/L2 MTL are the only methods that outperform the null model. With the DILGOM data none 
of the methods outperformed the null model when averaged over all target variables and when concentrating on the 
predictable target variables,  only the latent-noise BRRR and KRR with Gaussian kernel were able to outperform the 
null model. With gene expression prediction, latent noise BRRR (sample rank), GFlasso and the kernel methods
outperform the null model, latent-noise BRRR being the best. The econometrics data is the only case in which the 
independent-noise BRRR is more accurate  than the latent-noise BRRR on both metrics, 
and the latent-noise BRRR is the second best method. In this data set, however, the effects appear 
rather strong as different methods explain up to 10-29$\%$ of the variance of the target variables. 

On the small DILGOM data sets, L1/L2 MTL sets all regression weights to zero as hypothesized in Section 
\label{sec:selecting_var}. This demonstrates the need to develop new alternatives to L1/L2 regularization:
when modeling weak effects on small data sets, using L1/L2 penalties can prevent
analysis altogether. Regularization by making the noise model stronger as in latent-noise BRRR avoids this problem.

\begin{table}[ht]
\centering
\begin{tabular}{ccccc}
  \hline
 & econometrics & \begin{tabular}{c} DILGOM:\\gene expression \end{tabular} & \begin{tabular}{c} DILGOM:\\metabolomics \end{tabular} & fMRI \\ 
  \hline
\begin{tabular}{c} latent-noise \\ BRRR \end{tabular} & \bf{0.73320$\pm$0.22564\bf} & 
\bf{0.99990$\pm$0.00057\bf} & 1.00046$\pm$0.00130 & \bf{0.99798$\pm$0.00282\bf} \\[8mm] 
  \begin{tabular}{c} latent-noise \\ BRRR,\\sample rank \end{tabular} & \bf{0.73453$\pm$0.21219\bf} & 
1.00039$\pm$0.00107 & \bf{0.99995$\pm$0.00100\bf} &      \\[8mm] 
  \begin{tabular}{c} independent- \\ noise\\BRRR \end{tabular} & \bf{0.71072$\pm$0.20549\bf} & 1.00051$\pm$0.00038 & 1.04163$\pm$0.03781 & 1.00215$\pm$0.00183 \\[8mm] 
  L1/L2 MTL & \bf{0.75035$\pm$0.15651\bf} & 1.00000$\pm$0.00000 & 1.00000$\pm$0.00000 & 
\bf{0.99786$\pm$0.00090\bf} \\[8mm] 
  GFlasso &      & 1.00010$\pm$0.00106 & \bf{0.99996$\pm$0.00221\bf} &       \\[8mm] 
  \begin{tabular}{c} KRR with \\ linear \\ kernel \\ + PEER \end{tabular} & \bf{0.88138$\pm$0.11021\bf} & 1.00093$\pm$0.00057 & \bf{0.99995$\pm$0.00006\bf} & 1.00236$\pm$0.00112 \\[8mm] 
  \begin{tabular}{c} KRR with \\ Gaussian \\ kernel \\ + PEER \end{tabular} & \bf{0.90497$\pm$0.09707\bf} & \bf{0.99985$\pm$0.00016\bf} & \bf{0.99998$\pm$0.00004\bf} & 1.00649$\pm$0.00179 \\[8mm] 
  \begin{tabular}{c} BRRR \\ without \\ noise model \end{tabular} & \bf{0.81818$\pm$0.34747\bf} & 1.00568$\pm$0.00274 & 1.30795$\pm$0.08802 & 1.06722$\pm$0.05586 \\[8mm] 
  blm & 1.59040$\pm$1.23041 & 1.04245$\pm$0.00914 & 1.52573$\pm$0.08859 & 2.08650$\pm$0.25396 \\[8mm] 
  null model & 1.00000$\pm$0.00000 & 1.00000$\pm$0.00000 & 1.00000$\pm$0.00000 & 1.00000$\pm$0.00000 \\[8mm] 
   \hline
\end{tabular}
\caption{Test data MSE computed on the predictable target variables on the econometrics, DILGOM and fMRI data 
sets. Bold font indicates better than baseling accuracy.} 
\label{table:other_data_sets_MSE_predictable}
\end{table}

\subsection{Results: simultaneous modeling of both latent and independent structured noise} 
\label{sec:latent+independent_BRRR}
As both latent and independent structured noise can be present simultaneously, we evaluated the possible gains from
taking both noise types simultaneously into account. A model that incorporates both latent and independent 
structured noise, here called latent+independent-noise BRRR, was 
evaluated for the metabolomics prediction task on the NFBC1966 data and on the macroeconomic time series 
prediction task, the strong domains of the methods of interest. 

Results of this experiment are presented in Table \ref{table:latent_and_independent_noise_BRRR}. In metabolomics 
prediction, accounting for both noise types improved results slightly on the smallest training 
data size as compared to the best performing method latent-noise BRRR. On the larger 
training data sets, the more flexible latent+independent-noise BRRR model performed worse than the latent-noise 
BRRR that only accounts for latent noise. On the macroeconomic time series prediction task, accounting for both noise types improved performance 
as compared to only accounting for the dominant noise type (independent structured noise) on the smaller 
training data set. 
For summary,even though slight performance improvements were seen with the smallest training set sizes, the 
results indicate that as the size of the training data set increases, the advantages disappear. 
We hypothesize that the potential under-identifiability issues discussed in Section \ref{sec:latent_independent_differences} hinder model performance more than the increased 
flexibility improves it.

\begin{table}[ht]
\centering
\begin{tabular}{ccccc}
  \hline
 & \begin{tabular}{c} NFBC \\ $N = 3761$ \end{tabular} & \begin{tabular}{c} NFBC \\ $N = 1880$ \end{tabular} & 
\begin{tabular}{c} econometrics \\ $N = 120$ \end{tabular} & \begin{tabular}{c} econometrics \\ $N = 60$ 
\end{tabular} \\[8mm] 
  \hline
\begin{tabular}{c} latent-noise \\ BRRR \end{tabular} & \bf{0.9833$\pm$0.0077} & 0.9949$\pm$0.0037 & 
0.7536$\pm$0.2143 & 0.8374$\pm$0.1816 \\[8mm] 
  \begin{tabular}{c} latent+independent- \\ noise BRRR \end{tabular} & 0.9840$\pm$0.0072 & 
\bf{0.9947$\pm$0.0019} & 0.7445$\pm$0.1918 & \bf{0.7889$\pm$0.1561} \\[8mm] 
  \begin{tabular}{c} independent- \\ noise\\BRRR \end{tabular} & 0.9849$\pm$0.0078 & 0.9980$\pm$0.0059 & 
\bf{0.7339$\pm$0.1977} & 0.8097$\pm$0.2085 \\[8mm]
   \hline
\end{tabular}
\caption{Performance of the most flexible modeling assumptions. Test data MSE on the NFBC and econometrics data sets. 
On the larger training data sets, latent-noise BRRR and independent
noise BRRR outperform the model that accounts for both noise types,  latent+independent-noise BRRR. On the smaller 
training data sets, however, this model outperforms the models that only account for one noise type.} 
\label{table:latent_and_independent_noise_BRRR}
\end{table}

\subsection{Improvement in computational efficiency resulting from the reparameterization of model}\label{sec:runtime_experiment}
To confirm the computational speed-up resulting from the reparameterization presented in Section
\ref{sec:computation}, we performed an experiment where the algorithm implementing the na{\"i}ve Gibbs sampling
updates for the Bayesian reduced-rank regression \citep{Geweke96,karlsson2012conditional} was compared with
the new algorithm that uses the reparameterization. Similar improvements were achieved with all other BRRR models as well.

\begin{figure}[ht!]
\centering
\includegraphics[width=0.5\textwidth]{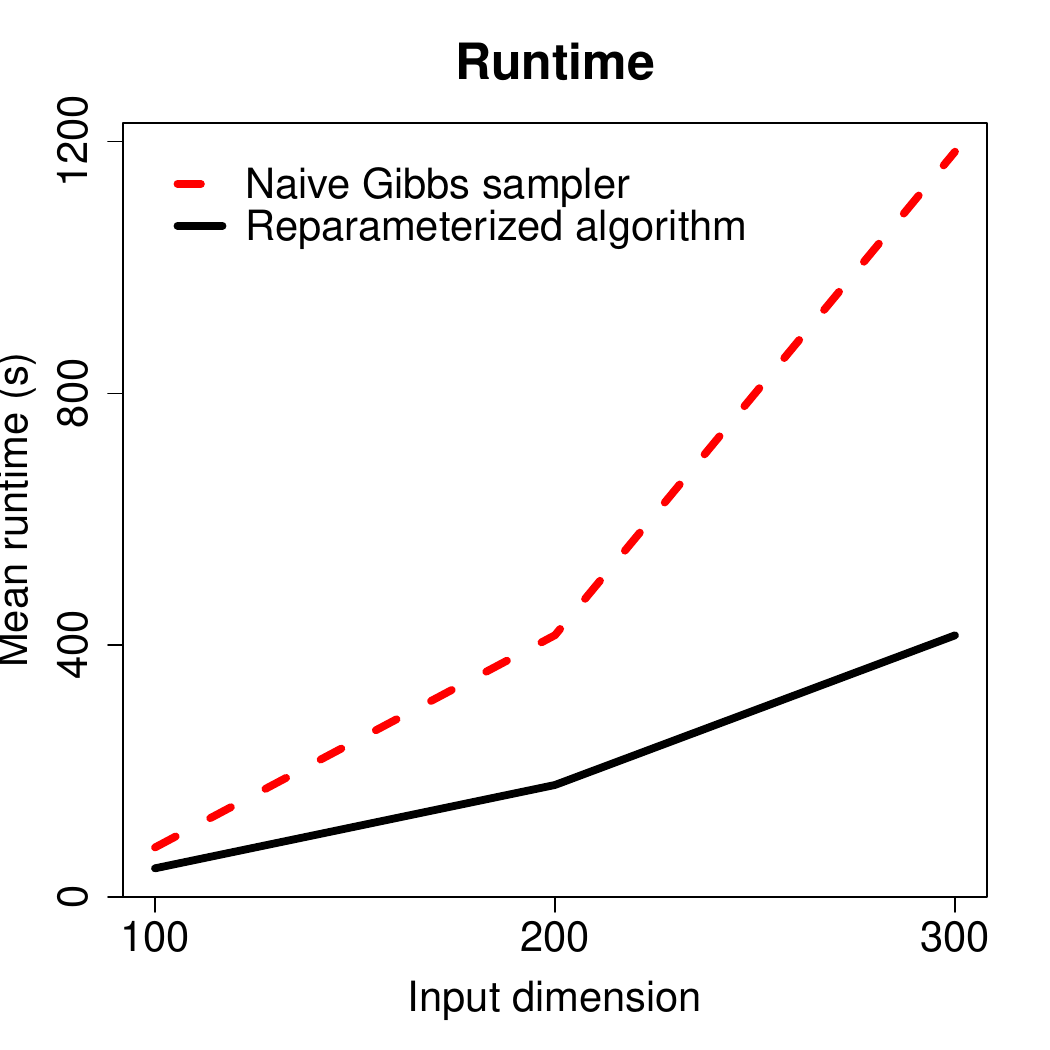} 
\caption{Runtime of the algorithm implementing the na{\"i}ve Gibbs sampler with computational complexity and the 
new algorith that
reparameterizes the model. The na{\"i}ve algorithm has a computational complexity of $O(P^3S_1^3)$ and the 
new algorithm $O(P^3 +S_1^3)$. Random variation over the repetitions was minimal and the error bars were 
omitted 
for clarity.}
\label{Fig_algorithm_runtime}
\end{figure}

Ten simulated data replicates were generated from the prior. The number of samples in the training set was fixed to
5000 and the number of target variables was set to 12. Rank of the regression coefficient matrix was 2.
Runtime was measured as a function of the number of covariates, which was varied from 100 to 300; 1000
posterior samples were generated.  The new 
algorithm that reparameterizes the model clearly outperformed the na{\"i}ve Gibbs sampler (Figure 
\ref{Fig_algorithm_runtime}). As a sanity check, the regression coefficient matrices estimated by the
 algorithms were compared , and found to be similar.

\subsection{Efficiency of the algorithm} \label{sec:efficiency_of_the_algorithm}
\begin{figure}[ht!]
\centering
 \includegraphics[width=0.8\textwidth]{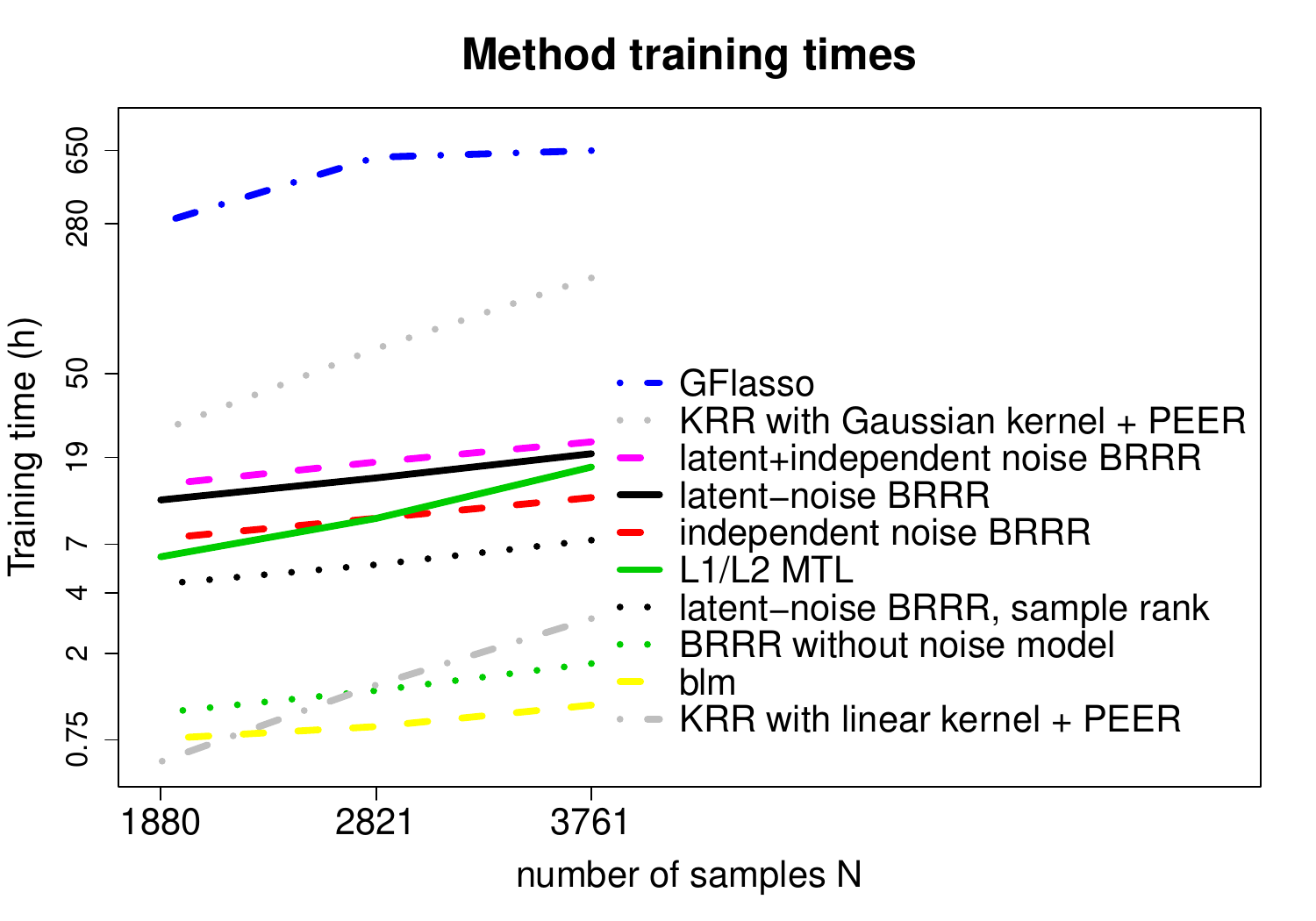}
\caption{Computation times of the methods for different training set sizes $N$ on the NFBC1966 metabolomics 
data.}
\label{Fig_NFBC66_ajoajat}
\end{figure}

To investigate the efficiency of the proposed algorithm and to compare it with the alternative methods, we recorded the 
wall-clock run times with the NFBC1966 data set, shown in Figure \ref{Fig_NFBC66_ajoajat}. In addition, we studied 
the conventional convergence diagnostics. To assess convergence and mixing, we re-computed four MCMC chains of 
2000 posterior samples each, for each of the BRRR methods. Averaged effective
sample sizes (ESS) and potential scale reduction factors (PSRF) were computed for 200 randomly selected parameters 
of the regression coefficient matrix \citep{gelman2004bayesian}. These results are presented in 
Table \ref{table:ESS_and_PSRF}.

All BRRR methods, except for independent-noise BRRR, converge (PSRF $<$ 1.1) and mix acceptably 
efficiently ($\frac{N_{\text{effective}}}{N_{\text{samples}}}\approx \frac{40}{1000}$ ). Independent noise 
BRRR, however, showed poor mixing and convergence. In initial experiments we observed that the PSRF for the
independent-noise BRRR did not necessarily ever reach values indicating convergence even when sampled for 15,000 
iterations. Thus, we decided to simply use the same number of MCMC iterations for each method in our experiments. 
The reason for the bad behaviour was the multimodality of the posterior distribution, caused by the too flexible 
model structure of the independent noise model, and the resulting convergence of the different chains into different 
modes.  

\begin{table}[ht]
\centering
\small
\begin{tabular}{ccccc}
  \hline
 & \begin{tabular}{c} independent noise \\ BRRR \end{tabular} & \begin{tabular}{c} BRRR without \\ noise model 
\end{tabular} & latent-noise BRRR & \begin{tabular}{c} latent-noise BRRR, \\ sample rank \end{tabular}
  \\ 
  \hline
1000 samples & 4.46 $\pm$ 0.32 & 1.03 $\pm$ 0.03 & 1.06 $\pm$ 0.05 & 1.01 $\pm$ 0.004 \\ 
  2000 samples & 3.42 $\pm$ 0.18 & 1.02 $\pm$ 0.02 & 1.05 $\pm$ 0.05 & 1.01 $\pm$ 0.003 \\ 
   \hline
\end{tabular}
\caption{Averaged PSRF} 
\label{table:ESS_and_PSRF}
\end{table}

\begin{table}[ht]
\centering
\begin{tabular}{ccccc}
    \hline
 & \begin{tabular}{c} independent noise \\ BRRR \end{tabular} & \begin{tabular}{c} BRRR without \\ noise 
model 
\end{tabular} & \begin{tabular}{c} latent-noise \\BRRR \end{tabular}
 & \begin{tabular}{c} latent-noise BRRR, \\ sample rank 
\end{tabular}
  \\ 
  \hline
1000 samples & 4.32 $\pm$ 0.32 & 43.88 $\pm$ 0.93 & 44.83 $\pm$ 0.25 & 40.74 $\pm$ 1.18 \\ 
  2000 samples & 5.15 $\pm$ 0.66 & 84.39 $\pm$ 1.61 & 86.40 $\pm$ 0.43 & 77.77 $\pm$ 1.25 \\ 
   \hline
\end{tabular}
\caption{Effective sample sizes for the Bayesian reduced rank regression 
methods. Independent-noise BRRR mixes substantially worse than the other methods.} 
\end{table}

\begin{figure}[ht!]
\centering
\begin{tabular}{cc}
. \includegraphics[width=0.45\textwidth]{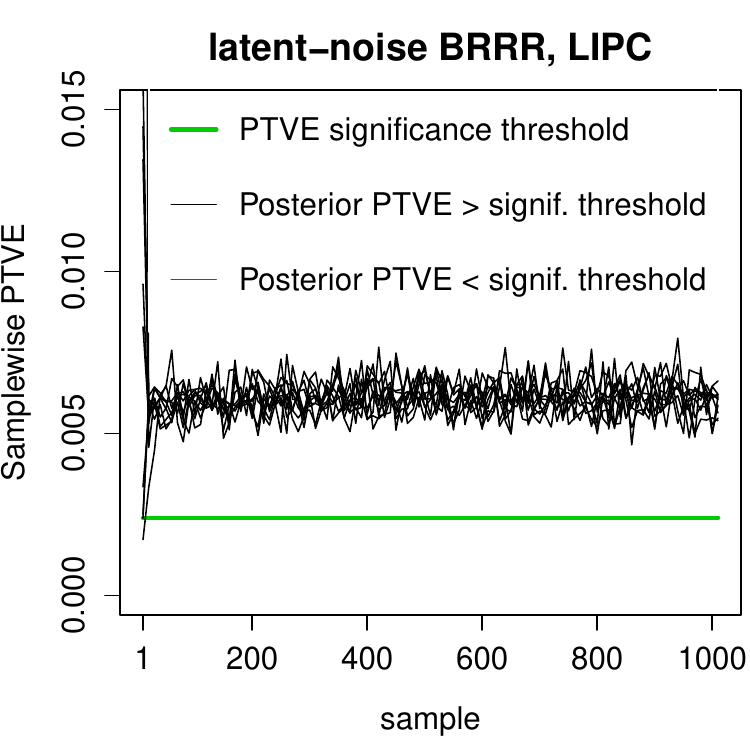} 
&
 \includegraphics[width=0.45\textwidth]{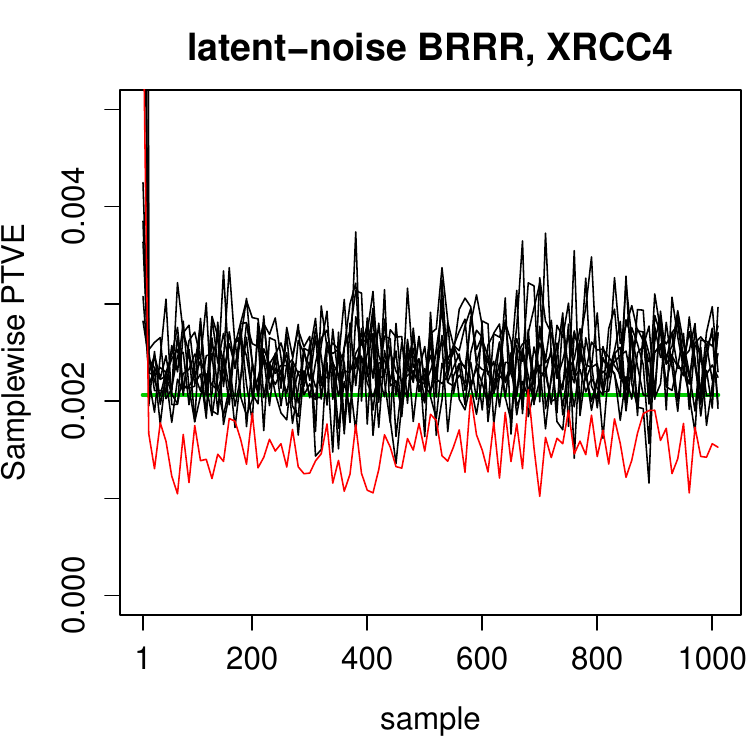} \\
(a) & (b) \\
. \includegraphics[width=0.45\textwidth]{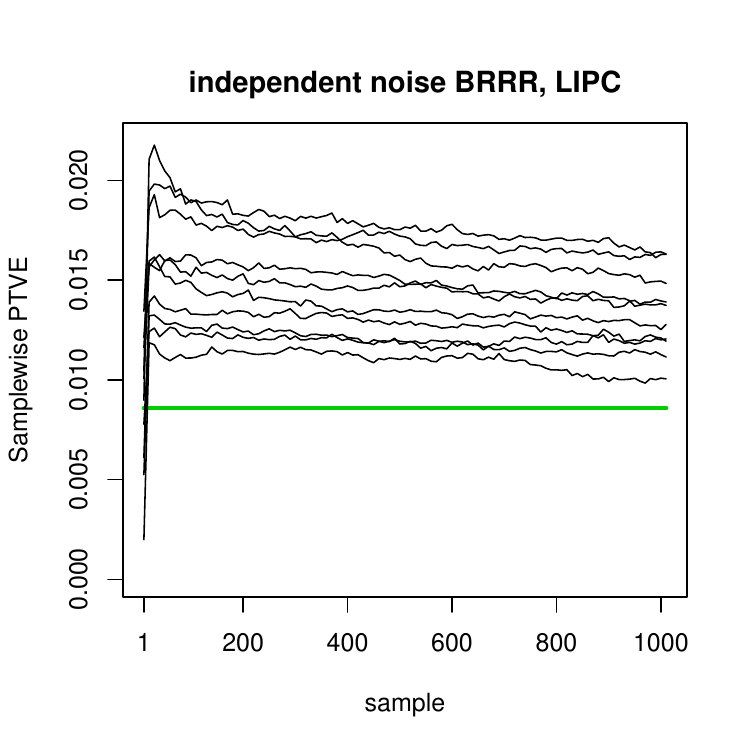} 
&
 \includegraphics[width=0.45\textwidth]{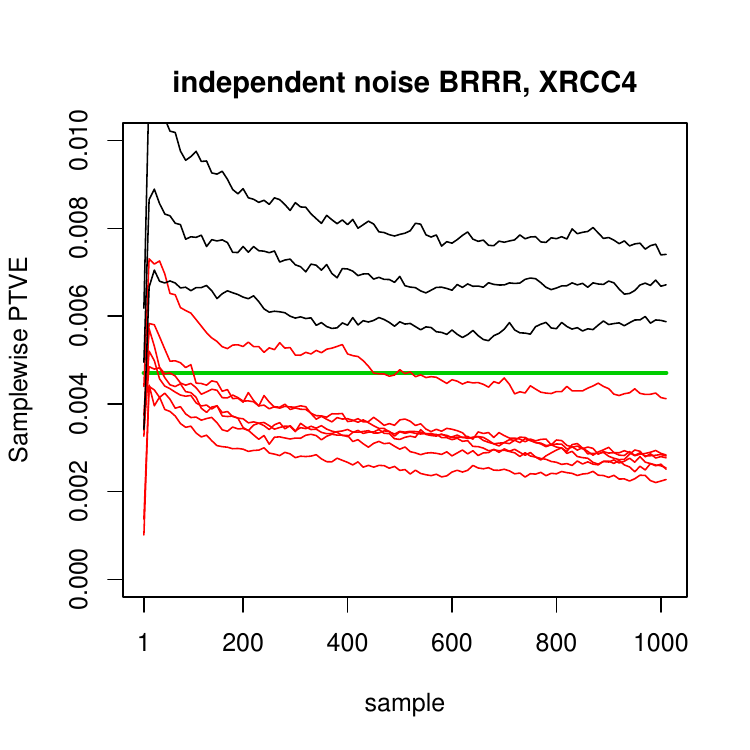} \\
(c) & (d)
\end{tabular}
\caption{Convergence plots of the association score parameter, that is, the proportion of the total variance explained (PTVE), 
for latent-noise BRRR and independent-noise BRRR. 10 MCMC chains were computed using data sets with 4702 
samples from genes \emph{LIPC} and \emph{XRCC4}. The green line marks the 0.05 significance level of the test score, 
obtained by permutation sampling. The chains, whose association scores exceed the significance threshold, are drawn
in black, whereas the chains that do not exceed it are drawn in red. Latent-noise BRRR converges and mixes 
appropriately: chains with different initializations converge and traverse the posterior. On the contrary, the 
independent-noise BRRR behaves rather pathologically: different chains converge to different solutions and 
explore the posterior poorly.}

\label{Fig_konvergenssi}
\end{figure}

To further demonstrate the difference between the latent-noise and independent-noise BRRR methods, we visualized the 
MCMC trace of the association metric used in Section \ref{sec:association_exp_res}. The instability of 
independent-noise BRRR is strikingly visible in Figure \ref{Fig_konvergenssi}. The chains converge to different modes 
and mix very slowly. On the other hand, the latent-noise BRRR appears to mix adequately and always converges to the 
same mode, except for one of the ten chains with the XRCC4 gene, which converges to a mode with a lower value of 
the explained variance.

\subsection{Results: summary of the results with the real data sets} \label{sec:overview}
To provide an overview of the performances of the different methods on the various data sets and tasks, 
the methods' performances were ranked for each task/data set. For the prediction tasks, methods were ranked 
according to the MSE on the test set. When none of the methods outperformed the null model, the scores on 
the predictable target variables were compared instead. In the association detection task, estimated 
statistical power was used as the ranking criterion. Table \ref{table:overview_MSE_Jussin_valikoima} presents 
the overview results.

Averaged over all data sets and tasks, latent-noise BRRR outperforms the comparison methods. In particular, the latent-noise 
BRRR outperforms the independent-noise BRRR on all setups except for the macroeconomic time series prediction 
task, where independent-noise BRRR is the best method and the two variants of latent-noise BRRR follow.
The difference between the latent-noise BRRR and the independent-noise BRRR is consistent, present on 4/5 test folds on the NFBC1966
metabolite prediction,  8/10 test folds on the DILGOM gene expression prediction, 10/10 test folds on DILGOM
metabolite prediction and on 2/2 folds on the fMRI response prediction.  On macroeconomic time series prediction,
independent-noise BRRR is better on 218/395 test folds. In the association detection task the latent-noise BRRR has higher power with both training set sizes on the challenging XRCC4 gene (0.94 vs. 0.32 with n=2,351; 0.98 vs. 0.41 with n=4,702).

Simultaneously accounting for both latent and independent structured noise improves performance on the smallest 
training data sets considered in the macroeconomic time series prediction and metabolomics prediction (NFBC1966) 
as compared to accounting for only one type of noise. On the other hand, with the larger training set sizes, the 
models with just the dominant noise type present perform better than the model including both noise types 
simultaneously. 

Selecting the rank for latent-noise BRRR by sampling or by cross-validation results in comparable 
performance. Average performance ranks for cross-validation based and sampling-based inferences are 2.5 and 
3.5, respectively. For the NFBC1966 data set and gene expression prediction task on the DILGOM data set, 
cross-validation yields better performance. On metabolomics prediction on DILGOM and the macroeconomic time 
series prediction, on the other hand, the sampling-based approach works better. It is also intriguing that similarly to  
the simulations, the sampling based variant of the model works better with independent structured 
noise (macroeconomic time series prediction) than the cross-validation based approach.

Latent-noise BRRR outperforms the null model on all test cases except for the metabolomics prediction on the 
DILGOM data. Even on that data set, however, the variant of the model that samples the maximum rank of the 
infinite prior outperforms the null model. We hypothesize that the poor performance may have resulted from 
convergence to some inferior mode of the posterior distribution; this can happen to latent-noise BRRR 
(as demonstrated in Figure \ref{Fig_konvergenssi}) although the sharing of information between the signal and noise models makes it substantially more stable than the
independent-noise BRRR.

\begin{table}[ht]
\centering
\begin{tabular}{c@{\hskip -6pt}c@{\hskip -6pt}c@{\hskip -6pt}c@{\hskip -6pt}c@{\hskip -6pt}c@{\hskip -6pt}c@{\hskip -6pt}c}
  \hline
 & \begin{tabular}{c} NFBC \\ $N = 3761$ \end{tabular} & \begin{tabular}{c} econometrics \\ $N = 120$ \end{tabular} & \begin{tabular}{c} DILGOM:\\gene \\ expression \\ $N = 458$ \end{tabular} & \begin{tabular}{c} DILGOM:\\metabolomics \\ $N = 458$ \end{tabular} & \begin{tabular}{c} fMRI \\ $N = 1307$ \end{tabular} & \begin{tabular}{c} NFBC: \\ XRCC4 \\ association \\ detection \\ $N = 4702$ \end{tabular} & \begin{tabular}{c} Average \\ rank \end{tabular} \\ 
  \hline
\begin{tabular}{c} latent-noise \\ BRRR \end{tabular} & \bf{1} &  2 &  2 &  7 &  2 & \bf{1} & \bf{2.5} \\[4mm] 
  \begin{tabular}{c} latent-noise \\ BRRR,\\sample rank \end{tabular} &  2 &  3 &  6 & \bf{1} &  &  & 3.0 \\[6mm] 
  \begin{tabular}{c} independent- \\ noise\\BRRR \end{tabular} &  3 & \bf{1} &  7 &  8 &  4 &  3 & 4.3 \\[6mm] 
  L1/L2 MTL &  7 &  4 &  4 &  5 & \bf{1} &  & 4.2 \\[4mm] 
  GFlasso &  4 &  &  5 &  3 &  &  & 4.0 \\[6mm] 
  \begin{tabular}{c} KRR with \\ linear \\ kernel \\ + PEER \end{tabular} &  6 &  6 &  8 &  2 &  5 &  & 5.4 \\[6mm] 
  \begin{tabular}{c} KRR with \\ Gaussian \\ kernel \\ + PEER \end{tabular} &  5 &  7 & \bf{1} &  4 &  6 &  & 4.6 \\[6mm] 
  \begin{tabular}{c} BRRR \\ without \\ noise model \end{tabular} &  8 &  5 &  9 &  9 &  7 &  & 7.6 \\[6mm] 
  blm & 10 &  9 & 10 & 10 &  8 &  & 9.4 \\[4mm] 
  null model &  9 &  8 &  3 &  6 &  3 &  & 5.8 \\[4mm] 
  cca &  &  &  &  &  &  4 &  \\[4mm] 
  lm &  &  &  &  &  &  2 &  \\[4mm] 
   \hline
\end{tabular}
\caption{Summary: ranking of methods according to performance in each studied data set and task.} 
\label{table:overview_MSE_Jussin_valikoima}
\end{table}

\section{Discussion}
In this work, we evaluated the performance of multiple-output regression with different assumptions for the 
structured noise. While most existing methods assume \emph{a priori} independence of the interesting effects 
and the uninteresting structured noise, we started from the opposite assumption of strong dependence between 
the components of the model. This assumption may be deemed appropriate  for instance with the molecular biological data 
sets often analyzed with such methods. Using simulations we demonstrated the harmfulness of the independence 
assumption when latent noise was present. In real data experiments the model assuming latent noise 
outperformed state-of-the-art methods in prediction of metabolite measurements from genotype (SNP) data and 
fMRI response prediction, and showed consistently good performance in the different domains. In an 
illustrative multivariate association detection task, the latent noise model had increased power to detect 
associations invisible to other methods. To better address the computational needs, we presented a new 
algorithm reducing the runtime considerably, and improving the scalability of the BRRR models as the number of 
variables increases.  The prior distributions were parameterized in terms of the new concept of \emph{latent 
signal-to-noise ratio}, which was a key ingredient for optimal model performance. In addition, the rotational 
unidentifiability of the model was solved using ordered infinite-dimensional shrinkage priors. We 
also demonstrated that the two modifications (model structure, regularization through the latent signal-to-noise ratio) 
made to the existing state-of-the-art noise modeling approach were both needed in order to reach the optimal performance.

In real data both latent and independent structured noise can be present. We studied a 
model incorporating both types simultaneously, and, based on these results, we concluded that the possible gains in 
predictive power as compared to modeling only the dominant type of noise were not worthwile. In fact, results 
were also found to degrade when both noise types were included, which we hypothesize to be the result of poor identifiability of the corresponding model

The new model implementing the concept of latent noise was studied using high-dimensional data containing weak 
signal (weak effects). The new model exploits a ubiquitous characteristic of such data: while the interesting 
effects are weak, the noise is strong. Latent-noise BRRR borrows statistical strength from the noise model so 
as to alleviate learning of the weak effects, by automatically enforcing the regression coefficients on 
correlated target variables to be correlated. This intuitive characteristic can be seen as a counterpart
of the powered correlation priors \citep{krishna2009bayesian} in the target variable space: Krishna et 
al.~used the correlation structure of the covariates as a prior for the regression weights to enforce
correlated covariates to have correlated weights.

The latent-noise BRRR is an extension of several common model families. By removing the covariates, the model 
reduces to a standard factor analysis model, which explains the output data with underlying factors. Thus, the 
latent-noise BRRR can be seen as a reversed analogy of PCA regression \citep{bernardo2003bayesian}, in which 
components of the input space are used as covariates in prediction; in latent-noise BRRR components derived 
from the output space are predicted using the covariates \citep[see][]{bo2009supervised}. Allowing the noise 
term to affect the latent space directly results in interesting connections to \emph{linear mixed models} 
(LMMs) and \emph{best linear unbiased   prediction} (BLUP) \citep{robinson1991blup}; using the latent noise
formulation, the model can explain away bias in the residuals as in BLUP. On the other hand, LMMs have a 
random term for each sample and target variable. While LMMs are not computationally feasible to generalize for 
high-dimensional targets due to the $NK$ random effect parameters and the associated inversion of an $NK 
\times NK$ covariance matrix, the latent-noise BRRR can be seen as a low-rank generalization of LMMs for 
high-dimensional target variables: the covariates are used for prediction in the latent space and in this
space there is a noise term for each sample and dimension. Therefore, the number of random effect parameters 
stays at $NS_1$ and inference remains tractable.

In summary, our findings extend the existing literature on modeling structured noise in an important way by 
showing that structured noise can, and should, be taken advantage of when learning the interesting effects 
between the covariates and the target variables, and how this can be done. Code in R for the new method is 
available for download at \url{http://users.ics.aalto.fi/lgillber/latent-noise/}\footnote{Code will be made available 
when the paper is published.}.


\acks{This work was financially supported by the Academy of Finland (the Finnish Centre of 
Excellence in Computational Inference Research COIN; grant numbers 259272 and 286607 to PM; grant number 257654 to
MP; grant number 140057 to SK).

NFBC1966 received financial support from the Academy of Finland (project grants 104781, 120315, 129269, 1114194, 24300796, Center of Excellence in Complex Disease Genetics and SALVE), University Hospital Oulu, Biocenter, University of Oulu, Finland (75617), NHLBI grant 5R01HL087679-02 through the STAMPEED program (1RL1MH083268-01), NIH/NIMH (5R01MH63706:02), ENGAGE project and grant agreement  HEALTH-F4-2007-201413, EU FP7 EurHEALTHAgeing -277849 and the Medical Research Council, UK (G0500539, G0600705, G1002319, 
PrevMetSyn/SALVE).

The development and applications of the
quantitative serum NMR metabolomics platform are supported by the Academy
of Finland, the Sigrid Juselius Foundation, Strategic Research Funding
from the University of Oulu, the British Heart Foundation, the Wellcome
Trust and the Medical Research Council, UK.

Disclosure: AJK, PS and MAK are shareholders of Brainshake Ltd., a company offering
NMR-based metabolite profiling.
}

\vskip 0.2in
\bibliography{bib_latent_noise}

\end{document}